\documentclass[11pt,oneside,onecolumn]{article}
\usepackage{yandex-research}
\usepackage{microtype}
\usepackage{amsmath,amssymb,booktabs,array}
\usepackage{tikz}
\usetikzlibrary{decorations.pathreplacing,positioning,calc}
\usepackage{longtable}

\usepackage{adjustbox}
\usepackage{cleveref}
\usepackage{xurl}

\title{FLAMIA: FLAsh MIcroscaling Attention for Fast and Causally Safe Training on Blackwell}
\author{Fedor Velikonivtsev, Mikhail Koltakov, Denis Kuznedelev}
\affiliations{Yandex Research}
\hypersetup{pdftitle={FLAMIA: FLAsh MIcroscaling Attention for Fast and Causally Safe Training on Blackwell},pdfauthor={Fedor Velikonivtsev, Mikhail Koltakov, Denis Kuznedelev}}
\newcommand{\codeurl}{https://github.com/yandex-research/flamia-blackwell-attention}

\begin{document}
\maketitle
\begin{abstract}
Attention remains a substantial computational cost in Transformer training, particularly at long context lengths. NVIDIA Blackwell GPUs provide high-throughput low-precision matrix multiplication, but exploiting it in attention requires addressing quantization overhead, causal information leakage through shared scales, and numerical inconsistencies between the forward and backward passes. We present FLAMIA, a family of attention kernels for Blackwell with MXFP8 and mixed NVFP4/FP8 training paths. Our design combines token-block scaling that reuses quantized operands across matrix orientations, targeted safeguards against scale-induced leakage, and consistent forward and backward computations. Fused quantization and architecture-specific kernel optimizations reduce overhead, while document-local scaling and workload-aware scheduling support variable-length documents directly in packed layouts without repacking. Across various causal packed-document workloads, the MXFP8 path accelerates the combined forward and backward attention call, including quantization, by factors of $1.10$--$1.40\times$ on B200 and $1.06$--$1.42\times$ on B300 relative to the fastest evaluated baseline for each workload. We validate training stability and quality preservation on pretraining Llama-style models. These results show how jointly designing quantization, training numerics, and document scheduling translates low-bit hardware throughput into practical attention acceleration.
\end{abstract}

\section{Introduction}

Softmax attention~\cite{vaswani2017attention} remains a substantial cost in Transformer training and inference, particularly at long context lengths. FlashAttention~\cite{dao2022flashattention} reduces memory traffic through tiled computation and online softmax. FlashAttention-4~\cite{zadouri2026fa4} further optimizes attention for NVIDIA Blackwell GPUs, but its BF16 implementation leaves higher low-precision throughput unexploited: Blackwell tensor cores support FP8 and FP4 matrix multiplication at twice and four times the BF16 rate, respectively~\cite{blackwelltuning}. Realizing these gains requires addressing quantization overhead and training numerics, while supporting the variable-length documents packed together during pretraining.

Recent 4-bit extensions of FlashAttention-4~\cite{hu2026hwfp4,fixedshift} motivate examining these requirements beyond single-step accuracy. Three constraints are central. First, hardware block scaling follows the contraction axis, whereas attention training uses operands along different contraction axes, potentially requiring separately quantized representations. Second, scales shared across positions can let future tokens affect earlier outputs despite a causal mask. Third, the backward pass must consistently reconstruct the forward's quantized intermediates: mismatches in quantization, tiling, or exponential evaluation can destabilize training despite high single-step cosine similarity. Packed inputs additionally require document-local scaling and load balancing. Among the evaluated implementations, some provide only a forward path for packed inputs~\cite{flashinfer}, use tensor-wide FP8 scaling~\cite{cudnn}, or lack causal-mask support~\cite{metalpfa4}.

We present FLAMIA (FLAsh MIcroscaling Attention), low-precision attention kernels for Blackwell with MXFP8 and mixed NVFP4/FP8 training paths. The mixed path uses NVFP4 for $\mathbf{Q}\mathbf{K}^{\top}$ and its recomputation, and 8-bit operands for the remaining matrix multiplications. Token-block scaling enables reuse of 8-bit representations across orientations. FLAMIA combines targeted leakage mitigations with consistent forward and backward computations and processes packed documents without repacking. Default scaling retains within-block dependencies and a sequence-wide backward accumulator dependency. Exact options remove them in supported batched causal calls. Our contributions are:
\begin{itemize}
\item Two training paths using token-block scaling to reduce quantized representation duplication across matrix orientations.
\item An analysis of scale-induced leakage and forward--backward inconsistencies, with targeted mitigations and training experiments showing that single-step accuracy proves neither causal correctness nor training stability.
\item Kernels combining fused quantization, coordinated softmax and matrix-multiplication scheduling, a low-precision two-CTA backward pass, and GPU-based scheduling for packed documents, plus two bit-level exponential approximations for inference.
\item An evaluation against FlashAttention-4, cuDNN, FlashInfer, Low Precision FA4~\cite{metalpfa4}, and the Hardware-Aware FP4 FlashAttention-4 (HW-FP4 FA4) of \citet{hu2026hwfp4}. Across all evaluated causal packed-document workloads, MXFP8 achieves the lowest combined forward and backward attention latency, including quantization: $1.10$--$1.40\times$ speedups on B200 and $1.06$--$1.42\times$ on B300 over the fastest evaluated baseline per workload. In 1B-model pretraining on 100B tokens, MXFP8 and mixed NVFP4/FP8 lead only to marginal increase of training loss and change of downstream metrics.
\end{itemize}

The code is available at \url{\codeurl}.

\section{Background and Related Work}

\paragraph{FlashAttention}
Let $\mathbf{Q}, \mathbf{K}, \mathbf{V} \in \mathbb{R}^{L \times d}$ be the query, key, and value tensors corresponding to a single head, with sequence length $L$ and head dimension $d$. The attention output is computed as follows:
\[
\begin{aligned}
\mathbf{S} &= \mathbf{Q}\mathbf{K}^{\top} / \sqrt{d}
    \in \mathbb{R}^{L \times L}, \\
\mathbf{P} &= \operatorname{softmax}(\mathbf{S})
    \in \mathbb{R}^{L \times L}, \\
\mathbf{O} &= \mathbf{P}\mathbf{V}
    \in \mathbb{R}^{L \times d}.
\end{aligned}
\]
The backward pass recomputes $\mathbf{P}$ and forms
\[
\begin{aligned}
\mathbf{dV} &= \mathbf{P}^{\top}\mathbf{dO}, & \mathbf{dP} &= \mathbf{dO}\,\mathbf{V}^{\top}, &
\mathbf{dS} &= \mathbf{P} \circ (\mathbf{dP} - D), \\
\mathbf{dQ} &= \mathbf{dS}\,\mathbf{K}, & \mathbf{dK} &= \mathbf{dS}^{\top}\mathbf{Q}, &&
\end{aligned}
\]
with the row sums $D_i = \sum_j P_{ij}\,dP_{ij}$, so every operand takes part in two products, contracted along different axes. A naive implementation materializes $\mathbf{P}$ and $\mathbf{S}$ in global device memory, resulting in a large memory footprint and substantial I/O traffic. FlashAttention \cite{dao2022flashattention} combines tiled computation with online softmax to keep intermediate tiles in fast on-chip memory. FlashAttention-2 \cite{dao2024flashattention2} reduces non-matmul operations and improves work partitioning, while FlashAttention-3 \cite{dao2024fa3} and FlashAttention-4 \cite{zadouri2026fa4} introduce optimizations tailored to NVIDIA Hopper and Blackwell GPUs, respectively. In our work, we focus on further improving attention performance on Blackwell, using BF16 FlashAttention-4 as a starting point.

\paragraph{Microscaling floating-point formats}
Microscaling formats \cite{rouhani2023mx} employ fine-grained quantization in which the elements of a block share a common scaling factor: each element is stored as a low-bit code, and its value is the code times the block's scaling factor. We call a scaling factor a \emph{scale} for short. NVIDIA Blackwell provides native hardware support for these formats \cite{blackwelltuning}. MXFP formats (MXFP4, MXFP6, and MXFP8) use blocks of 32 elements with UE8M0 scaling factors, whereas NVFP4 uses blocks of 16 elements with E4M3 scaling factors and an additional FP32 tensor-level scaling factor \cite{nvidia2025nvfp4}. Blackwell's block-scaled matrix multiply (MMA) takes such operands directly and applies their scales as it accumulates. Throughout the paper, we write $\hat{\mathbf{X}}$ for the quantized form of a tensor $\mathbf{X}$: its low-bit codes, 8-bit E4M3 or 4-bit E2M1, together with one scaling factor per block of codes. These formats have been applied to post-training quantization in FP-Quant \cite{egiazarian2026fpquant} and to low-precision training \cite{tseng2025mxfp4,castro2025quartet,quartet2,nvidia2025nvfp4,chmiel2025fp4}. These training methods primarily target GEMMs in linear layers. Full-Stack FP4 \cite{fullstackfp4} also explores mixed-precision attention.

\paragraph{Low-precision attention}
The SageAttention series \cite{zhang2024sage,zhang2024sage2,zhang2025sage2pp,zhang2025sage3} accelerates self-attention by quantizing matmul operands and mitigating quantization error to preserve model quality. SageAttention3 \cite{zhang2025sage3} targets microscaling FP4 for inference and separately explores 8-bit forward and backward passes for training. More recent work includes trainable INT8 attention in SageBwd \cite{sagebwd2026}, a mixed FP4/FP8 training path in HW-FP4 FA4 \cite{hu2026hwfp4}, and MXFP8 forward and backward kernels in Low Precision FA4 \cite{metalpfa4}. Attn-QAT \cite{attnqat} emphasizes matching backward score recomputation to the quantized forward pass and correcting precision assumptions in the gradient calculation. Complementarily, Qiu and Yao \cite{qiu2025lowprec} analyze training instability caused by biased rounding errors and low-rank attention representations. In our work, we develop an optimized implementation of low-precision attention aimed at improving hardware utilization while addressing the numerical consistency and causal correctness issues discussed in the following section.

\section{Hardware-Aware Design Around Low-Bit Attention Constraints}
\label{sec:method}


Blackwell tensor cores provide higher peak throughput for FP8 and FP4 than for BF16. Exploiting this throughput in attention requires coordinating three aspects of the design: scaling layouts compatible with the hardware, control of scale-induced dependencies across token positions, and numerical consistency between the forward and backward passes. We first describe these constraints and the resulting MXFP8 and mixed NVFP4/FP8 training paths, then present the kernel optimizations, packed-document support, and inference-only approximation

\subsection{Token-Block Scaling}
\label{sec:scales}

Block-scaled matrix multiply--accumulate (MMA) instructions require scaling groups along each operand's contraction axis. The 4-bit instructions used here also require K-major layouts. Attention training complicates this requirement because each of its six logical operands participates in two products with different contraction axes. $\mathbf{Q}$ and $\mathbf{K}$ are contracted along the head dimension in $\mathbf{S}$, but along queries and keys in $\mathbf{dK}$ and $\mathbf{dQ}$, respectively. $\mathbf{V}$ and $\mathbf{dO}$ are contracted along the head dimension in $\mathbf{dP}$, but along keys and queries in $\mathbf{P}\mathbf{V}$ and $\mathbf{dV}$. Likewise, $\mathbf{P}$ is contracted along keys in $\mathbf{P}\mathbf{V}$ and queries in $\mathbf{dV}$, while $\mathbf{dS}$ is contracted along keys in $\mathbf{dQ}$ and queries in $\mathbf{dK}$. Conventional one-dimensional block scaling therefore generally requires two quantized representations per operand, one per contraction axis. Each representation requires writing its codes and scales, and each separate quantization pass rereads the source tensor. Representations of $\mathbf{Q}$, $\mathbf{K}$, and $\mathbf{V}$ needed for backward computation must remain in memory after the forward pass. Because quantization is memory-bound, this duplication across six operands can consume a substantial fraction of attention training time.

We reduce this overhead through selective NVFP4 computation and reusable E4M3 representations. In the mixed NVFP4/FP8 path, $\mathbf{Q}$ and $\mathbf{K}$ use NVFP4 with one scale per 16 consecutive head-dimension elements of each token. The same representations serve forward score computation and backward score recomputation, which share a contraction axis. The remaining products use E4M3 operands, including separate representations of $\mathbf{Q}$ and $\mathbf{K}$ for gradient computation. Our \emph{token-block} scaling shares one power-of-two exponent across 32 consecutive tokens and the full head dimension. $\mathbf{dS}$ instead shares one exponent per $32 \times 32$ block, as \Cref{sec:numerics} discusses. Repeating these exponents across hardware scaling groups yields MXFP8-compatible representations. The common exponent within each $32 \times d$ token block preserves the validity of its codes and scales under transposition. Square $\mathbf{dS}$ blocks give the same property for contractions along queries and along keys. Since 8-bit MMA supports both K-major and transposed (MN-major) layouts, each E4M3 payload serves both orientations without a separately quantized transposed copy.

\subsection{Causally Safe Scales}
\label{sec:causal}

Causal masking alone does not prevent information leakage through data-dependent quantization scales. If a future token increases the maximum of a shared scale group, the resulting scale change can alter the reconstructed values of earlier tokens. An earlier output can then depend on a masked future input through the quantizer.

\begin{figure}[!htbp]
\centering
\def\lmfont{\scriptsize}
\def\lmpad{1.2}
\def\lmtile{7.5}
\def\lmkey{4.2}
\def\lmsplit{7}
\setbox0\hbox{\lmfont keys $\rightarrow$}\edef\lmhead{\the\dimexpr\ht0+\dp0\relax}%
\begin{tikzpicture}[x=1mm,y=1mm,font=\lmfont,
  every node/.style={inner sep=0pt,outer sep=0pt},
  node distance=\lmpad mm,
  off/.style={fill=blue!12,draw=black!45,line width=0.3pt},
  dia/.style={fill=orange!35,draw=black!45,line width=0.3pt},
  skip/.style={fill=white,draw=black!20,line width=0.3pt},
  past/.style={fill=blue!12,draw=black!45,line width=0.3pt},
  fut/.style={fill=red!45,draw=black!45,line width=0.3pt},
  futq/.style={fill=black!8,draw=black!45,line width=0.3pt}]
\pgfmathsetmacro{\lmgrid}{4*\lmtile}
\begin{scope}[yshift=-\lmhead-\lmpad mm]
\foreach \r in {0,...,3}{
  \foreach \c in {0,...,3}{
    \pgfmathtruncatemacro{\d}{\c-\r}
    \ifnum\d<0 \path[off] (\c*\lmtile,-\r*\lmtile) rectangle ++(\lmtile,-\lmtile);\fi
    \ifnum\d=0 \path[dia] (\c*\lmtile,-\r*\lmtile) rectangle ++(\lmtile,-\lmtile);\fi
    \ifnum\d>0 \path[skip] (\c*\lmtile,-\r*\lmtile) rectangle ++(\lmtile,-\lmtile);\fi
  }}
\coordinate (gridNW) at (0,0);
\coordinate (gridSW) at (0,-\lmgrid);
\coordinate (gridNE) at (\lmgrid,0);
\draw[black!70,line width=0.5pt] (gridNW) -- (\lmgrid,-\lmgrid);
\node[above=of $(gridNW)!0.5!(gridNE)$] {keys $\rightarrow$};
\node[rotate=90,above=of $(gridNW)!0.5!(gridSW)$] {$\leftarrow$ queries};
\node[anchor=north east,align=right] at ($(gridNE)+(-\lmpad,-\lmpad)$) {masked,\\skipped};
\draw[black,line width=0.7pt] (2*\lmtile,-2*\lmtile) rectangle ++(\lmtile,-\lmtile);
\end{scope}
\begin{scope}[shift={(\lmgrid+\lmsplit,0)}]
  \coordinate (row) at (0,-\lmhead-2*\lmpad mm);
  \foreach \k in {0,...,4}{\path[past] ($(row)+(\k*\lmkey,0)$) rectangle ++(\lmkey,-\lmkey);}
  \path[fut] ($(row)+(5*\lmkey,0)$) rectangle ++(\lmkey,-\lmkey);
  \foreach \k in {6,7}{\path[futq] ($(row)+(\k*\lmkey,0)$) rectangle ++(\lmkey,-\lmkey);}
  \draw[black,line width=0.8pt,dashed] ($(row)+(5*\lmkey,\lmpad)$) -- ++(0,-\lmkey-2*\lmpad);
  \node[above=of {$(row)+(5*\lmkey,\lmpad)$}] {causal boundary of $i$};
  \draw[decorate,decoration={brace,amplitude=1.5mm,mirror}]
    ($(row)+(0,-\lmkey-\lmpad)$) -- ($(row)+(8*\lmkey,-\lmkey-\lmpad)$) coordinate[midway] (bmid);
  \node[below=1.5mm+\lmpad mm of bmid,align=center,anchor=north] (glabel)
    {one scale group of $\mathbf{V}$ along keys\\shared scale $s = \max_j |\mathbf{V}_j| / 6$};
  \coordinate (legy) at ($(glabel.south)+(0,-2*\lmpad)$);
  \path[off] (row |- legy) rectangle ++(3,-3) coordinate (sw1);
  \node[anchor=west] at ($(sw1)+(\lmpad,1.5)$) {grouped $\mathbf{V}$ scales};
  \path[dia] ($(sw1)+(-3,-\lmpad)$) rectangle ++(3,-3) coordinate (sw2);
  \node[anchor=west] at ($(sw2)+(\lmpad,1.5)$) {key-local $\mathbf{V}$ scales};
\end{scope}
\begin{scope}[yshift=-\lmhead-\lmpad mm]
\draw[black!60,densely dotted] (3*\lmtile,-2*\lmtile) -- (row);
\draw[black!60,densely dotted] (3*\lmtile,-3*\lmtile) -- ($(row)+(0,-\lmkey)$);
\end{scope}
\end{tikzpicture}
\caption{Left: under a causal mask, the boundary between visible and masked keys runs through the diagonal tiles. Right: the keys that query $i$ reads in the marked tile. A scale group of $\mathbf{V}$ spans keys on both sides of the boundary, so the future key (red) can raise the group maximum, grow the shared scale $s$ and re-round the codes of the past keys. On diagonal tiles we give each key its own scale and keep the cheaper grouped scales elsewhere.}
\label{fig:leak}
\end{figure}
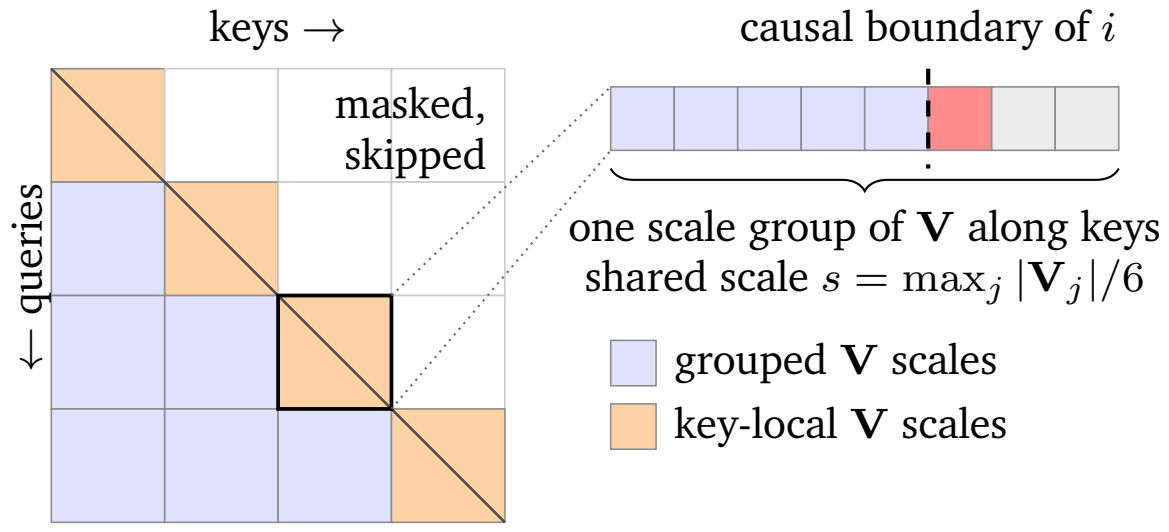

\Cref{fig:leak} shows where this happens. FlashAttention computes $\mathbf{S}$ in tiles of $B_r$ queries by $B_c$ keys, with $B_r = B_c = 128$ in our kernels. Under a causal mask, the boundary between visible and masked keys runs through the \emph{diagonal} tiles (left panel). In $\mathbf{P}\mathbf{V}$ the contraction runs over keys, so a scale group of $\mathbf{V}$ spans several keys. Inside a diagonal tile, such a group can hold keys on both sides of a query's boundary (right panel). With a 4-bit $\mathbf{V}$, a larger group scale changes the codes of the visible keys.

With an 8-bit $\mathbf{V}$, the same rescale is usually harmless. When the group scale doubles, each code keeps its mantissa and only its exponent drops by one, so the value it represents stays the same. This fails in two cases: an element that drops below the normal range of E4M3 loses mantissa bits, and an element that was clamped at the largest code is no longer clamped. The second case occurs in the reference microscaling conversion of \citet{rouhani2023mx}. It sets the shared exponent to $\lfloor \log_2 \max_i |v_i| \rfloor - e_{\max}$, so the scaled group maximum can exceed 448, the largest E4M3 value, and is clamped. A later key that raises the maximum doubles the scale, the earlier element is no longer clamped, and its code changes. Whole-tensor statistics, such as a per-tensor maximum, are the extreme case: through them, a later token can change every earlier output.

Our scaling removes these leaks or confines them to one block. No operand scale covers more than 32 consecutive positions: token blocks share one exponent across 32 tokens, and $\mathbf{Q}$ and $\mathbf{K}$ use per-token NVFP4 scales, which cannot leak at all. A later token can therefore change earlier results only inside its own block of 32, through elements near the bottom of the normal range. We also round each scale's exponent up, so the scaled maximum never exceeds the largest code and no element is clamped.

When $\mathbf{V}$ is 4-bit, we give it a per-key power-of-two scale and fold this scale into $\hat{\mathbf{P}}$. A per-key scale depends on a single key and cannot leak. Only diagonal tiles contain a group that straddles the causal boundary, so we use per-key scales on diagonal tiles and keep the cheaper grouped scales elsewhere. The backward has two more leaks on the key side: $\mathbf{V}$ in $\mathbf{dP}$, and $\mathbf{K}$ in $\mathbf{dQ} = \mathbf{dS}\,\mathbf{K}$. We close both the same way, by folding a per-key power-of-two scale into the other operand. We call the version for $\mathbf{K}$ the $\mathbf{K}$ fold. Appendix~\labelcref{app:design} describes the small residue that remains inside a block and the \emph{exact options} that, together with the causal $\mathbf{dQ}$ scale below, remove it.

The backward has one more leak, through the FP16 $\mathbf{dQ}$ accumulator. As \Cref{sec:twocta} explains, we reduce the $\mathbf{dQ}$ partials in FP16, whose five exponent bits leave little range, so we multiply the partials by a power of two that keeps them in range. By default this factor is sized from maxima over the whole sequence, so a later token can change how every earlier row of $\mathbf{dQ}$ rounds. We therefore also provide a causal $\mathbf{dQ}$ scale, which sizes the factor of each query row only from the keys that the row can see. 

Validation loss measured through the training kernel alone is insufficient to detect leakage. In our controlled experiments, a model trained with a leaking kernel obtains lower loss when evaluated through that kernel but performs worse under exact causal attention. We therefore compare both evaluation modes in the leakage experiments of \Cref{sec:eval_numerics}.

\subsection{Forward--Backward Numerical Consistency}
\label{sec:numerics}

FlashAttention recomputes attention probabilities during the backward pass from $\mathbf{Q}$, $\mathbf{K}$, and saved row statistics rather than storing the full probability matrix. It also computes $D_i=\mathbf{dO}_i\cdot\mathbf{O}_i$ from the saved output. In exact arithmetic, this equals $\sum_j P_{ij}\,dP_{ij}$ when $\mathbf{O}=\mathbf{P}\mathbf{V}$ and $\mathbf{dP}=\mathbf{dO}\mathbf{V}^{\top}$. Since the rows of $\mathbf{P}$ sum to one, each row of the softmax gradient $\mathbf{dS}=\mathbf{P}\circ(\mathbf{dP}-D)$ then sums to zero. Quantization can introduce additional inconsistency if these computations use different reconstructed operands. Matching those operands removes one discrepancy but does not by itself restore the exact identities: quantized probabilities need not sum to one, and output rounding and accumulation introduce further error.

We therefore align the representations and numerical conventions used by the two passes. Score recomputation uses the same quantized $\mathbf{Q}$ and $\mathbf{K}$ representations as the forward pass for the selected precision path. The backward pass also reproduces the relevant tiling and diagonal decisions, exponential evaluation, and scaling relative to the row maximum. We compute $D$ using the same quantized $\mathbf{dO}$ used in $\mathbf{dP}$, with an output formed from the corresponding E4M3 value representation. These choices reduce representation-induced discrepancies without eliminating ordinary rounding or normalization error.

Our experiments show that mismatched tiling or exponential evaluation can destabilize training even when single-step gradient similarity remains high. Such mismatches can introduce structured errors whose effects accumulate across training steps, as \citet{barron2026fixsoftmaxcrossentropy} also find when the forward and backward passes evaluate a softmax differently. A single-step comparison therefore cannot establish training stability. Building on the quantization consistency emphasized by Attn-QAT~\cite{attnqat}, we also examine these sources of mismatch in \Cref{sec:eval_numerics}.

Before computing $\mathbf{dQ}$ and $\mathbf{dK}$, we quantize $\mathbf{dS}$ to E4M3. Its entries span a wide dynamic range, with many substantially smaller than nearby maxima. An excessively large scale increases the fraction of small entries rounded to zero, even when their contribution to the norm error is small. In our ablations, this fraction tracks training degradation more closely than the reported norm error, and the scale choice has the largest effect on loss among the numerical choices examined. We therefore derive a scale from the maximum absolute value in each $32\times32$ block, using the tile already held by the kernel. The square scaling regions permit reuse of one quantized payload in both $\mathbf{dQ}$, which contracts over keys, and $\mathbf{dK}$, which contracts over queries. Appendix~\labelcref{app:design} describes the FP16 $\mathbf{dQ}$ accumulator and rounding choices.

\subsection{MXFP8 and Mixed NVFP4/FP8 Paths}
\label{sec:scheme}

Although our library supports an all-4-bit training path, it is not the default: in our evaluation, it is slower than paths retaining most operands in 8 bits and reaches a slightly higher loss. The 4-bit MMA instructions used here have no MN-major form, and each orientation needs its own copy. The six quantized copies make quantization the largest part of this path's backward time. A 4-bit $\mathbf{V}$ adds two costs. Its key-axis scale groups require per-key scaling on diagonal tiles (\Cref{sec:causal}), increasing work in the forward pass. Moreover, $\mathbf{P}\mathbf{V}$ and $\mathbf{dP}$ contract $\mathbf{V}$ along different axes, requiring separate 4-bit representations, while the computation of $D$ must remain consistent with the representation used in $\mathbf{dP}$. One E4M3 token-block representation can serve both products. Finally, the evaluated backward products are not tensor-core-bound, limiting the benefit of narrower operands. \Cref{sec:eval_ablation} quantifies these costs.

Our default paths use MXFP8 or NVFP4 for score computation and recomputation. The mixed NVFP4/FP8 path quantizes $\mathbf{Q}$ and $\mathbf{K}$ with one NVFP4 scale per 16 head-dimension elements of each token. The MXFP8 path uses E4M3 with token-block scales. Both paths use E4M3 token-block representations of $\mathbf{V}$, $\mathbf{dO}$, and the $\mathbf{Q}$ and $\mathbf{K}$ operands of the gradient products, E4M3 $\mathbf{dS}$ with one scale per $32 \times 32$ block, and E4M3 probability operands. Returned outputs and gradients are BF16, while $\mathbf{dQ}$ partials are accumulated in FP16. We describe the scaling of $\hat{\mathbf{P}}$ and the kernels below.

Where enabled, a normalized Hadamard transform, as in the incoherence processing of QuIP~\cite{chee2024quip} and QuIP\#~\cite{Tseng2024QuIPEB}, mixes the head channels of both $\mathbf{Q}$ and $\mathbf{K}$ before quantization. The shared orthogonal transform preserves their dot product in exact arithmetic while spreading channel outliers. Rounding and quantization can still change the resulting scores. It is enabled by default at head dimension 128 on the two default paths. Appendix~\labelcref{app:design} reports its effects and remaining format choices.

For fixed-scale E4M3 probability quantization, a multiplier $2^c$ reduces underflow. Conditional rescaling retains the online-softmax row maximum until the new maximum exceeds it by $\tau$. In the base-2 representation, the unnormalized weights relative to the retained maximum can reach $2^{\tau}$, or $2^{c+\tau}$ after scaling. Since the largest finite E4M3 value is 448, we enforce $c+\tau\leq\log_2 448$. Violating this bound can saturate the weights and corrupt the output without a NaN or runtime error. Appendix~\labelcref{app:kernels} gives the parameters and probability-quantization details.

\subsection{Kernel Design}
\label{sec:kernels}

Our kernels extend the warp-specialized FlashAttention-4 design of \citet{zadouri2026fa4}. A load warp, an MMA warp, two softmax warpgroups, correction warps, and an epilogue warp exchange $\mathbf{S}$, $\hat{\mathbf{P}}$, and $\mathbf{O}$ through tensor memory (TMEM). \Cref{sec:eval_ablation} evaluates these modifications.

\subsubsection{Quantization and Token-Block Scales}

Each attention invocation quantizes its current operands. For each input tensor, we fuse the required Hadamard rotation, block statistics, and generation of codes and scales for all required representations into one launch. The $\mathbf{dO}$ launch additionally computes $D$ and initializes the $\mathbf{dQ}$ accumulator. Fusion avoids input rereads during quantization preprocessing. Outputs and returned gradients remain in BF16 and require no output quantization scales, while $\mathbf{dS}$ scales are computed within the attention kernel. The required scales are therefore derived from current tensors without a separate output-calibration pass.

Token-block scales are constant along the head dimension and factor algebraically out of the score dot product. Writing $q_Q,q_K$ for stored codes and $e_Q,e_K$ for their shared exponents, the reconstructed score is $\widetilde S_{ij}=2^{e_Q(i)+e_K(j)}\sum_k q_{Q,ik}q_{K,jk}/\sqrt{d}$. The MXFP8 path therefore uses an unscaled E4M3 MMA, folds the row factor into the softmax scale, and applies the column factors separately. This preserves the scaling algebraically. Bitwise equality also depends on range and accumulation order.

\subsubsection{Softmax Throughput and MMA Scheduling}

Exponential evaluation is a major throughput bottleneck in the evaluated forward kernels. A $128 \times 128$ tile requires 16384 exponentials, corresponding to 1024 cycles at B200's MUFU throughput, so faster low-bit MMAs alone do not remove this bottleneck. We address it in three ways. First, as in FA4, FlashInfer, and Low Precision FA4, we evaluate a fraction of the exponentials using a clamped polynomial on the FMA pipeline, selecting the fraction per path and GPU. B300 doubles the exponential throughput of B200. There, polynomial emulation becomes overhead in the pure-E4M3 configuration, and we disable it. Second, the NVFP4 path quantizes probability operands in the log domain. Since $\max_j 2^{x_j}=2^{\max_j x_j}$, we compute each 32-key group's maximum before exponentiation and subtract its scaling exponent from the exponential argument. This eliminates a per-entry division and a subsequent scaling pass, producing the largest individual NVFP4 forward improvement in \Cref{sec:eval_ablation}. Third, on B300, \texttt{tcgen05.ld.red} combines the row-maximum reduction with the TMEM score load. We use it on tiles known to be fully unmasked, including off-diagonal tiles in causal calls.

We overlap the remaining work with matrix multiplication. The MMA warp splits the next tile's $\mathbf{Q}\mathbf{K}^{\top}$ into two 64-key operations, issued before and after the current tile's $\hat{\mathbf{P}}\mathbf{V}$, respectively. This keeps the tensor cores active while the softmax warps process the current tile. In the backward pass, a checkerboard assignment distributes emulated exponentials so that warps sharing an SM sub-partition use different execution pipelines.

\subsubsection{Low-Bit 2-CTA Backward}
\label{sec:twocta}

FlashAttention-4's two-CTA backward mode uses a cluster of two cooperative thread arrays (CTAs) to share MMA operands and cover 256 keys over a query range. The evaluated FA4 implementation provides this mode in BF16. In its accumulator configuration, all 512 TMEM columns are occupied when $\mathbf{dP}$ is issued, leaving no capacity for additional block-scale operands. Token-block scaling permits the compute warps to apply these factors outside the MMA, enabling a low-precision implementation without TMEM scale storage for $\mathbf{dP}$.

Our implementation supports MHA and GQA on batched and packed inputs. On packed inputs, paired key tiles may belong to different documents, and each CTA masks positions outside its document. As in \citet{metalpfa4}, FP16 $\mathbf{dQ}$ partials are written through bulk reduce-add, halving reduction traffic relative to FP32. FP16's five exponent bits limit its range: in our tests, the fixed scaling factor used by Low Precision FA4 overflowed for outlier $\mathbf{K}$ channels. We instead use a data-dependent power-of-two prescale, with the dependency qualifications in \Cref{sec:causal}. An empirical rule per shape and format selects the two-CTA mode.

\subsection{Packed-Document Attention}
\label{sec:varlen}

\subsubsection{Packed Layout}

Pretraining packs documents contiguously into rows of $T$ tokens. Low-precision attention on these rows faces two challenges. First, shared scales can couple otherwise independent documents: a document with large activations can coarsen the representation of others sharing its scale. Second, document starts need not align with the fixed $128 \times 4$ scale-factor atoms consumed by the block-scaled MMA. The evaluated cuDNN FP8 path uses tensor-wide scales, while its MXFP8 path requires a scale layout padded per document. Padding documents to aligned starts resolves alignment but copies $\mathbf{Q}$, $\mathbf{K}$, $\mathbf{V}$, and gradients in each layer.

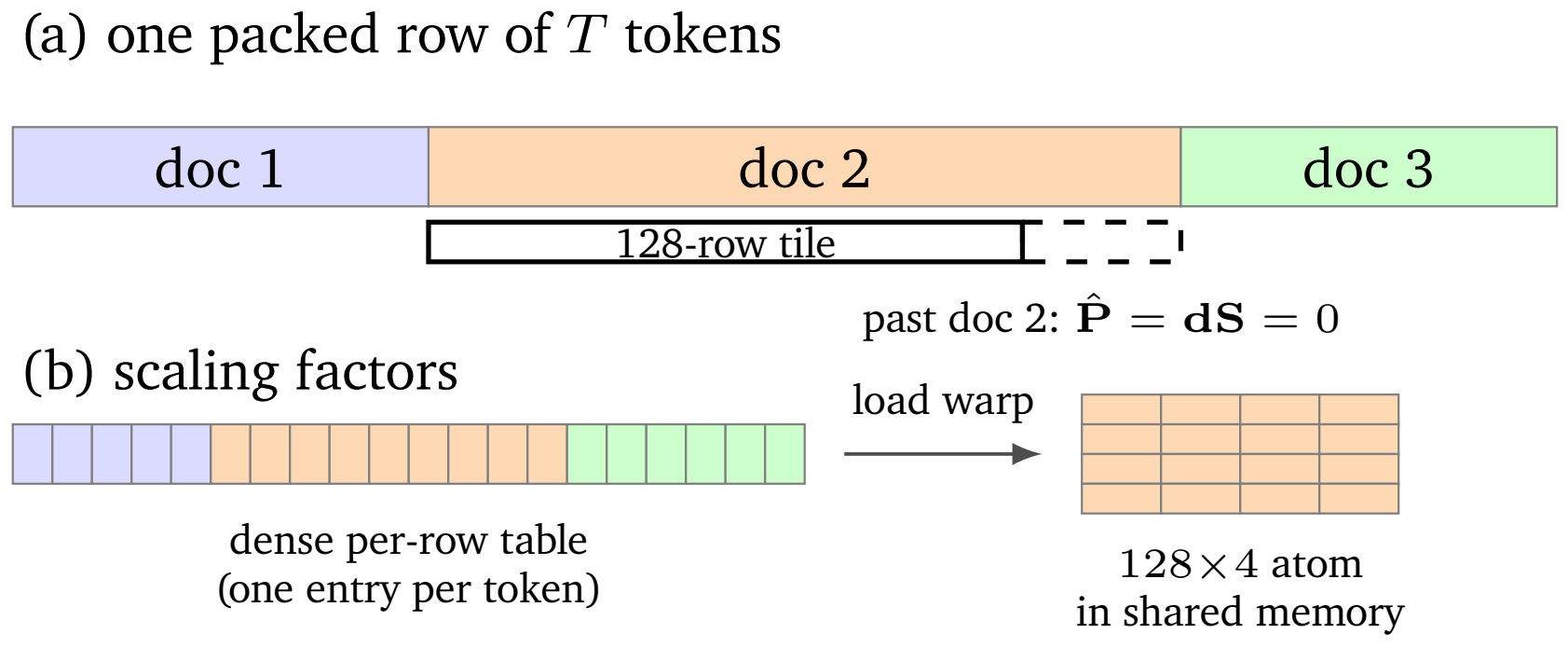
\begin{figure}[!htbp]
\centering
\resizebox{0.75\linewidth}{!}{
\begin{tikzpicture}[x=1mm,y=1mm,font=\scriptsize,
  docA/.style={fill=blue!14,draw=black!50,line width=0.3pt},
  docB/.style={fill=orange!30,draw=black!50,line width=0.3pt},
  docC/.style={fill=green!20,draw=black!50,line width=0.3pt},
  tbl/.style={draw=black!50,line width=0.3pt},
  arr/.style={-{latex},line width=0.5pt,black!70}]
\node[anchor=west] at (-1,4.5) {(a) one packed row of $T$ tokens};
\path[docA] (0,0) rectangle (21,-4);   \node at (10.5,-2) {doc 1};
\path[docB] (21,0) rectangle (59,-4);  \node at (40,-2) {doc 2};
\path[docC] (59,0) rectangle (78,-4);  \node at (68.5,-2) {doc 3};
\draw[black,line width=0.7pt] (21,-4.8) rectangle (51,-6.8);
\node[font=\tiny] at (36,-5.8) {128-row tile};
\draw[black,line width=0.7pt,dashed] (51,-4.8) rectangle (59,-6.8);
\node[anchor=north,font=\tiny] at (55,-7.0) {past doc 2: $\hat{\mathbf{P}}=\mathbf{dS}=0$};
\node[anchor=west] at (-1,-12.5) {(b) scaling factors};
\foreach \i in {0,...,19}{
  \pgfmathsetmacro{\c}{\i<5 ? "blue!14" : (\i<14 ? "orange!30" : "green!20")}
  \path[tbl,fill=\c] (\i*2,-15) rectangle ++(2,-3);}
\node[anchor=north,font=\tiny,align=center] at (20,-18.6) {dense per-row table\\(one entry per token)};
\draw[arr] (42,-16.5) -- (52,-16.5);
\node[anchor=south,font=\tiny] at (47,-16.3) {load warp};
\foreach \r in {0,...,3}{\foreach \c in {0,...,3}{\path[tbl,fill=orange!30] (54+\c*4,-13.5-\r*1.5) rectangle ++(4,-1.5);}}
\node[anchor=north,font=\tiny,align=center] at (62,-20.0) {$128{\times}4$ atom\\in shared memory};
\end{tikzpicture}}

\caption{Documents packed contiguously. (a) A 128-row tile may start inside a document. (b) The load warp assembles the tile's scale-factor atom in shared memory from dense per-token tables.}
\label{fig:packed}
\end{figure}

Our quantizers process $\mathbf{Q}$, $\mathbf{K}$, $\mathbf{V}$, and $\mathbf{dO}$ directly in the packed layout, keeping operand token blocks and their statistics within document boundaries. They write dense scale metadata with one entry per token. This operand-local construction does not remove the default accumulator-scale dependency discussed in \Cref{sec:causal}. Tile boundaries depend on document lengths and offsets (\Cref{fig:packed}a), so the load warp assembles each tile's $128 \times 4$ scale-factor atom in shared memory from the per-token tables (panel b). On the all-4-bit path, two keys share a byte. For odd document offsets we keep a second copy of $\mathbf{V}^{\top}$ shifted by one key, which Appendix~\labelcref{app:kernels} describes.

Keys outside the active document are masked, and invalid query rows receive an infinite log-sum-exp sentinel so that their probability and gradient contributions vanish. Epilogues write outputs and gradients directly to the packed buffers. The layout preserves the intended per-document computation. Bitwise equality of gradients also depends on accumulation order, because the default $\mathbf{dQ}$ reduction uses atomics. Buffer capacities depend on $T$, allowing different document-length distributions to reuse storage and the same captured CUDA graph.

\subsubsection{Load-Balanced Scheduling}

Variable document lengths produce substantial differences in tile cost. In the evaluated FA4 schedule, tiles are ordered longest-first within each sequence, but sequences are visited in input order. A long document near the end can therefore delay completion. The evaluated FlashInfer path instead constructs a load-balancing plan on the CPU and transfers it to the GPU for each new workload.

We construct the schedule on the GPU once per microbatch without host synchronization. Each work item contains 256 consecutive queries from one document, corresponding to a CTA's two 128-query tiles. Its estimated cost is the number of key blocks attended to, determined by document length and, for causal attention, query position. Items across the row are sorted by decreasing cost. A persistent grid assigns them to CTAs in alternating directions, distributing long and short items across the grid. Highly skewed workloads can leave CTAs idle under static assignment. Causal calls and non-causal calls whose longest document is at least four times the mean therefore use Cluster Launch Control (CLC) to acquire items dynamically, as FA4 does for causal calls. Requests follow the current item's final $\mathbf{K}$ and $\mathbf{V}$ loads, avoiding premature work reservation.

In the backward pass, each CTA processes a key tile across query tiles. We similarly order key tiles by decreasing estimated work and group heads into sections for L2 locality, as FA4's batched schedule does. Appendix~\labelcref{app:kernels} describes list construction and the backward launch layout.

\subsection{Bit-Level Exponentials for Inference}
\label{sec:inference}

Since the exponential limits the forward, we also offer two cheaper approximations of it for inference. Both build $2^{x}$ directly in the bits of a float. The bit pattern of a positive float, read as an integer, grows almost linearly with $\log_2$ of its value. Writing $x$ in fixed point into the exponent field therefore gives the piecewise-linear approximation of \citet{mitchell1962computer}, $2^{x} \approx 2^{\lfloor x \rfloor}(1 + \{x\})$.

A 4-bit $\hat{\mathbf{P}}$ tolerates an inexact exponential. E2M1 has only eight magnitudes, at least 33\,\% apart, while Mitchell's approximation is off by a few percent, so the approximate $2^{x}$ almost always falls into the same rounding interval as the exact value and gives the same code. Our \emph{Mitchell-P} quantizer therefore computes $2^{x}$ with ordinary arithmetic instructions instead of the hardware exponential unit (MUFU) that limits the forward. A fused multiply-add writes $x$ in fixed point into the bits of a float, an integer shift moves these bits into the exponent and mantissa, and a hardware convert rounds the resulting approximation of $2^{x}$ to E2M1. A constant offset of $x$ centres the approximation error, and an integer clamp keeps the result valid for any score range. Like the Direct-P quantizer of HW-FP4 FA4~\cite{hu2026hwfp4}, it maps scores to codes without an exact exponential, but where Direct-P fits an affine classifier, Mitchell-P reproduces almost every code of exact rounding. The EFQ-Softmax of \citet{efqsoftmax} also produces codes without computing an exponential.

An 8-bit $\hat{\mathbf{P}}$ needs a more accurate value. Neighbouring E4M3 values lie only 7--12.5\,\% apart, close to the error of the approximation, so the error would change many codes and shift $\hat{\mathbf{P}}$ instead of vanishing in the rounding. For this format we use the variant of \citet{schraudolph1999fast}, which computes the same $2^{\lfloor x \rfloor}(1 + \{x\})$ in BF16 and offsets $x$ so that the error averages zero. A fused multiply-add adds to $2^{7}x$ a constant that fixes the FP32 exponent and adds the BF16 exponent bias. The low 16 bits then hold the BF16 pattern of the approximation, with the integer part of $x$ in its exponent and 7 fraction bits in its mantissa, and a 16-bit shift turns them into an FP32 value. We use it for five of every eight exponential pairs and keep the rest exact on MUFU.

We use neither approximation in training. As \Cref{sec:numerics} explains, the backward must rebuild the forward's $\hat{\mathbf{P}}$ exactly, and a backward with the same approximation lost more time than the forward gained. \Cref{sec:eval_inference} measures both approximations, and Appendix~\labelcref{app:infexp} derives them.
\section{Experiments}
\label{sec:experiments}

We evaluate latency, accuracy, optimizations, inference approximations, and training behavior, benchmarking kernels on B200 and B300. Baselines include BF16 and FP8 FlashAttention-4~\cite{zadouri2026fa4}; cuDNN BF16, FP8, and MXFP8 attention~\cite{cudnn}; FlashInfer~\cite{flashinfer}; Low Precision FA4~\cite{metalpfa4}; HW-FP4 FA4~\cite{hu2026hwfp4}, with its Direct-P forward and its causal path; and PyTorch SDPA on nested tensors. FlashInfer configurations include FP8 CUTLASS and CuTe-DSL backends and NVFP4 KV-cache prefill. Baselines use their supported ragged APIs. Low Precision FA4 lacks causal masking, and neither it nor the evaluated nested-tensor SDPA path supports GQA. cuDNN MXFP8, FA4 FP8, FlashInfer, and HW-FP4 FA4 lack a supported packed-document backward path in this evaluation and appear only in forward comparisons. cuDNN MXFP8 and Low Precision FA4 use $1\times32$ scaling blocks; Low Precision FA4 retains both orientations. Appendix~\labelcref{app:hardware} specifies hardware, software, and timing.

\begin{table}[!htbp]
\centering
\caption{Training call on packed documents (forward and backward, quantization included), causal, 16 heads, $d = 128$: time in ms (fastest in bold), TFLOP/s in small type; Speedup: the fastest baseline's time over ours.}\label{tab:varlen_train}
\scriptsize
\setlength{\tabcolsep}{2.5pt}
\renewcommand{\arraystretch}{1.0}
\begin{adjustbox}{max width=\linewidth}
\begin{tabular}{@{}lllllllll@{}}
\toprule
& & \multicolumn{2}{l}{\textbf{Ours}} & \textbf{cuDNN} & \textbf{FA4} & \textbf{cuDNN} & \textbf{SDPA} & \\
\cmidrule(lr){3-4}
\textbf{GPU} & \textbf{Documents} & MXFP8 & NVFP4 & FP8 & BF16 & BF16 & BF16 & \textbf{Speedup} \\
\midrule
B200 & $4{\times}4096$ & \textbf{0.89}\,{\tiny 1079} & 0.92\,{\tiny 1049} & 1.06\,{\tiny 912} & 1.08\,{\tiny 889} & 1.28\,{\tiny 751} & 6.30\,{\tiny 153} & 1.18$\times$ \\
 & $16{\times}4096$ & \textbf{3.53}\,{\tiny 1090} & 3.64\,{\tiny 1058} & 3.90\,{\tiny 986} & 4.24\,{\tiny 907} & 4.88\,{\tiny 790} & 13.4\,{\tiny 288} & 1.10$\times$ \\
 & $16{\times}1024$ & \textbf{0.44}\,{\tiny 542} & 0.49\,{\tiny 492} & 0.53\,{\tiny 456} & 0.50\,{\tiny 478} & 0.55\,{\tiny 440} & 5.94\,{\tiny 41} & 1.14$\times$ \\
 & $64{\times}1024$ & \textbf{1.66}\,{\tiny 579} & 1.82\,{\tiny 530} & 1.85\,{\tiny 521} & 2.00\,{\tiny 482} & 2.25\,{\tiny 427} & 5.79\,{\tiny 166} & 1.11$\times$ \\
 & $8192 + 32{\times}256$ & \textbf{0.91}\,{\tiny 1095} & 0.94\,{\tiny 1059} & 1.25\,{\tiny 793} & 1.10\,{\tiny 898} & 1.55\,{\tiny 639} & 5.36\,{\tiny 185} & 1.22$\times$ \\
 & $8192 + 224{\times}256$ & \textbf{1.80}\,{\tiny 650} & 1.99\,{\tiny 589} & 3.03\,{\tiny 387} & 2.52\,{\tiny 466} & 6.71\,{\tiny 175} & 7.62\,{\tiny 154} & 1.40$\times$ \\
 & log-unif.\ 16k & \textbf{0.71}\,{\tiny 885} & 0.75\,{\tiny 838} & 0.89\,{\tiny 706} & 0.85\,{\tiny 738} & 1.00\,{\tiny 631} & 6.31\,{\tiny 100} & 1.20$\times$ \\
 & log-unif.\ 16k, GQA & \textbf{0.71}\,{\tiny 881} & 0.73\,{\tiny 860} & 0.99\,{\tiny 637} & 0.85\,{\tiny 740} & 1.02\,{\tiny 618} & {\tiny\texttt{no GQA}} & 1.19$\times$ \\
 & log-unif.\ 64k & \textbf{3.61}\,{\tiny 1061} & 3.70\,{\tiny 1034} & 4.15\,{\tiny 922} & 4.32\,{\tiny 886} & 5.12\,{\tiny 748} & 13.8\,{\tiny 277} & 1.15$\times$ \\
\midrule
B300 & $4{\times}4096$ & \textbf{0.79}\,{\tiny 1220} & 0.84\,{\tiny 1141} & 0.90\,{\tiny 1065} & 0.99\,{\tiny 968} & 1.15\,{\tiny 840} & 8.87\,{\tiny 99} & 1.14$\times$ \\
 & $16{\times}4096$ & \textbf{3.17}\,{\tiny 1212} & 3.28\,{\tiny 1175} & 3.35\,{\tiny 1147} & 3.99\,{\tiny 965} & 4.45\,{\tiny 865} & 13.4\,{\tiny 289} & 1.06$\times$ \\
 & $16{\times}1024$ & \textbf{0.41}\,{\tiny 588} & 0.45\,{\tiny 532} & 0.48\,{\tiny 507} & 0.47\,{\tiny 518} & 0.51\,{\tiny 469} & 9.41\,{\tiny 26} & 1.14$\times$ \\
 & $64{\times}1024$ & \textbf{1.52}\,{\tiny 635} & 1.68\,{\tiny 573} & 1.67\,{\tiny 578} & 1.83\,{\tiny 527} & 2.08\,{\tiny 462} & 8.62\,{\tiny 112} & 1.10$\times$ \\
 & $8192 + 32{\times}256$ & \textbf{0.82}\,{\tiny 1210} & 0.87\,{\tiny 1146} & 1.07\,{\tiny 926} & 1.04\,{\tiny 954} & 1.41\,{\tiny 703} & 8.04\,{\tiny 114} & 1.27$\times$ \\
 & $8192 + 224{\times}256$ & \textbf{1.67}\,{\tiny 705} & 1.84\,{\tiny 636} & 2.81\,{\tiny 418} & 2.37\,{\tiny 495} & 6.47\,{\tiny 181} & 9.45\,{\tiny 124} & 1.42$\times$ \\
 & log-unif.\ 16k & \textbf{0.64}\,{\tiny 985} & 0.69\,{\tiny 911} & 0.77\,{\tiny 814} & 0.78\,{\tiny 810} & 0.91\,{\tiny 696} & 8.62\,{\tiny 73} & 1.21$\times$ \\
 & log-unif.\ 16k, GQA & \textbf{0.67}\,{\tiny 943} & 0.68\,{\tiny 925} & 0.87\,{\tiny 728} & 0.78\,{\tiny 810} & 0.93\,{\tiny 680} & {\tiny\texttt{no GQA}} & 1.16$\times$ \\
 & log-unif.\ 64k & \textbf{3.23}\,{\tiny 1186} & 3.35\,{\tiny 1144} & 3.60\,{\tiny 1063} & 4.11\,{\tiny 932} & 4.70\,{\tiny 816} & 14.2\,{\tiny 264} & 1.11$\times$ \\
\bottomrule
\end{tabular}
\end{adjustbox}
\end{table}

\begin{table}[!htbp]
\centering
\caption{Forward kernel on packed documents, causal, 16 heads, $d = 128$: time in ms (fastest in bold), TFLOP/s in small type; Speedup: the fastest baseline's time over ours on B200 GPU. cuDNN MX: cuDNN MXFP8; FlashInfer: its fastest backend.}
\label{tab:varlen_fwd}
\scriptsize
\setlength{\tabcolsep}{2pt}
\renewcommand{\arraystretch}{1.05}
\begin{adjustbox}{max width=0.98\linewidth}
\begin{tabular}{@{}llllllllllllll@{}}
\toprule
& & \multicolumn{2}{l}{\textbf{Ours}} & \multicolumn{4}{l}{\textbf{8-bit}} & \textbf{4-bit} & \multicolumn{4}{l}{\textbf{BF16}} & \\
\cmidrule(lr){3-4}\cmidrule(lr){5-8}\cmidrule(lr){9-9}\cmidrule(lr){10-13}
\textbf{GPU} & \textbf{Documents} & MXFP8 & NVFP4 & \shortstack{cuDNN\\FP8} & \shortstack{cuDNN\\MX} & \shortstack{FA4\\FP8} & \shortstack{Flash\\Infer} & \shortstack{Flash\\Infer} & FA4 & cuDNN & \shortstack{Flash\\Infer} & SDPA & \textbf{Speedup} \\
\midrule
B200 & $4{\times}4096$ & \shortstack{\textbf{0.17}\\{\tiny 1616}} & \shortstack{0.21\\{\tiny 1293}} & \shortstack{0.24\\{\tiny 1137}} & \shortstack{0.40\\{\tiny 682}} & \shortstack{0.27\\{\tiny 1018}} & \shortstack{0.25\\{\tiny 1104}} & \shortstack{0.24\\{\tiny 1138}} & \shortstack{0.24\\{\tiny 1153}} & \shortstack{0.42\\{\tiny 658}} & \shortstack{0.28\\{\tiny 993}} & \shortstack{1.99\\{\tiny 138}} & 1.40$\times$ \\
 & $16{\times}4096$ & \shortstack{\textbf{0.70}\\{\tiny 1569}} & \shortstack{0.82\\{\tiny 1340}} & \shortstack{0.75\\{\tiny 1468}} & \shortstack{1.50\\{\tiny 733}} & \shortstack{1.06\\{\tiny 1038}} & \shortstack{0.91\\{\tiny 1211}} & \shortstack{0.85\\{\tiny 1300}} & \shortstack{0.98\\{\tiny 1121}} & \shortstack{1.50\\{\tiny 733}} & \shortstack{1.15\\{\tiny 956}} & \shortstack{3.96\\{\tiny 278}} & 1.07$\times$ \\
 & $16{\times}1024$ & \shortstack{\textbf{0.07}\\{\tiny 1011}} & \shortstack{0.07\\{\tiny 916}} & \shortstack{0.11\\{\tiny 621}} & \shortstack{0.26\\{\tiny 264}} & \shortstack{0.12\\{\tiny 593}} & \shortstack{0.15\\{\tiny 455}} & \shortstack{0.10\\{\tiny 715}} & \shortstack{0.10\\{\tiny 652}} & \shortstack{0.22\\{\tiny 308}} & \shortstack{0.16\\{\tiny 433}} & \shortstack{1.92\\{\tiny 36}} & 1.41$\times$ \\
 & $64{\times}1024$ & \shortstack{\textbf{0.24}\\{\tiny 1158}} & \shortstack{0.27\\{\tiny 1035}} & \shortstack{0.34\\{\tiny 804}} & \shortstack{1.15\\{\tiny 240}} & \shortstack{0.45\\{\tiny 612}} & \shortstack{0.53\\{\tiny 515}} & \shortstack{0.34\\{\tiny 811}} & \shortstack{0.41\\{\tiny 665}} & \shortstack{1.05\\{\tiny 262}} & \shortstack{0.61\\{\tiny 448}} & \shortstack{2.06\\{\tiny 134}} & 1.42$\times$ \\
 & $8192 + 32{\times}256$ & \shortstack{\textbf{0.18}\\{\tiny 1598}} & \shortstack{0.22\\{\tiny 1279}} & \shortstack{0.30\\{\tiny 957}} & \shortstack{0.42\\{\tiny 667}} & \shortstack{0.28\\{\tiny 1023}} & \shortstack{0.37\\{\tiny 772}} & \shortstack{0.29\\{\tiny 978}} & \shortstack{0.24\\{\tiny 1172}} & \shortstack{0.44\\{\tiny 644}} & \shortstack{0.42\\{\tiny 681}} & \shortstack{1.92\\{\tiny 147}} & 1.37$\times$ \\
 & $8192 + 224{\times}256$ & \shortstack{\textbf{0.28}\\{\tiny 1204}} & \shortstack{0.32\\{\tiny 1045}} & \shortstack{0.48\\{\tiny 698}} & \shortstack{3.70\\{\tiny 91}} & \shortstack{0.59\\{\tiny 564}} & \shortstack{0.67\\{\tiny 500}} & \shortstack{0.84\\{\tiny 397}} & \shortstack{0.55\\{\tiny 613}} & \shortstack{3.61\\{\tiny 93}} & \shortstack{0.77\\{\tiny 438}} & \shortstack{2.51\\{\tiny 134}} & 1.72$\times$ \\
 & log-unif.\ 16k & \shortstack{\textbf{0.13}\\{\tiny 1360}} & \shortstack{0.16\\{\tiny 1112}} & \shortstack{0.20\\{\tiny 890}} & \shortstack{0.35\\{\tiny 509}} & \shortstack{0.22\\{\tiny 831}} & \shortstack{0.23\\{\tiny 799}} & \shortstack{0.19\\{\tiny 969}} & \shortstack{0.18\\{\tiny 977}} & \shortstack{0.34\\{\tiny 537}} & \shortstack{0.24\\{\tiny 736}} & \shortstack{2.02\\{\tiny 89}} & 1.39$\times$ \\
 & log-unif.\ 16k, GQA & \shortstack{\textbf{0.13}\\{\tiny 1357}} & \shortstack{0.16\\{\tiny 1111}} & \shortstack{0.20\\{\tiny 886}} & \shortstack{0.34\\{\tiny 526}} & \shortstack{0.16\\{\tiny 1147}} & \shortstack{0.32\\{\tiny 570}} & \shortstack{0.18\\{\tiny 976}} & \shortstack{0.16\\{\tiny 1095}} & \shortstack{0.33\\{\tiny 546}} & \shortstack{0.38\\{\tiny 469}} & {\tiny\texttt{no GQA}} & 1.18$\times$ \\
 & log-unif.\ 64k & \shortstack{\textbf{0.73}\\{\tiny 1505}} & \shortstack{0.86\\{\tiny 1269}} & \shortstack{0.82\\{\tiny 1341}} & \shortstack{1.68\\{\tiny 650}} & \shortstack{1.10\\{\tiny 995}} & \shortstack{1.01\\{\tiny 1087}} & \shortstack{0.91\\{\tiny 1204}} & \shortstack{1.03\\{\tiny 1063}} & \shortstack{1.65\\{\tiny 665}} & \shortstack{1.22\\{\tiny 894}} & \shortstack{4.15\\{\tiny 263}} & 1.12$\times$ \\
\bottomrule
\end{tabular}
\end{adjustbox}
\end{table}

\begin{table}[!htbp]
\centering
\caption{B200. Backward kernels on packed documents, causal, 16 heads, $d = 128$: time in ms (fastest in bold), TFLOP/s in small type; Speedup: the fastest baseline's time over ours.}\label{tab:varlen_bwd}
\scriptsize
\setlength{\tabcolsep}{1.5pt}
\renewcommand{\arraystretch}{1.0}
\begin{adjustbox}{max width=\linewidth,max totalheight=0.78\textheight}
\begin{tabular}{@{}lllllllll@{}}
\toprule
& & \multicolumn{2}{l}{\textbf{Ours}} & \textbf{FP8} & \multicolumn{3}{l}{\textbf{BF16}} & \\
\cmidrule(lr){3-4}\cmidrule(lr){6-8}
\textbf{GPU} & \textbf{Documents} & MXFP8 & NVFP4 & cuDNN & FA4 & cuDNN & SDPA & \textbf{Speedup} \\
\midrule
B200 & $4{\times}4096$ & 0.52\,{\tiny 1312} & \textbf{0.51}\,{\tiny 1354} & 0.69\,{\tiny 993} & 0.73\,{\tiny 944} & 0.79\,{\tiny 867} & 2.56\,{\tiny 269} & 1.36$\times$ \\
 & $16{\times}4096$ & 2.14\,{\tiny 1286} & \textbf{2.08}\,{\tiny 1324} & 2.65\,{\tiny 1039} & 3.09\,{\tiny 890} & 3.31\,{\tiny 831} & 9.20\,{\tiny 299} & 1.27$\times$ \\
 & $16{\times}1024$ & 0.23\,{\tiny 748} & \textbf{0.23}\,{\tiny 755} & 0.30\,{\tiny 572} & 0.37\,{\tiny 467} & 0.32\,{\tiny 543} & 2.56\,{\tiny 67} & 1.32$\times$ \\
 & $64{\times}1024$ & 0.88\,{\tiny 779} & \textbf{0.88}\,{\tiny 785} & 1.07\,{\tiny 642} & 1.47\,{\tiny 468} & 1.24\,{\tiny 553} & 3.04\,{\tiny 226} & 1.22$\times$ \\
 & $8192 + 32{\times}256$ & 0.53\,{\tiny 1333} & \textbf{0.52}\,{\tiny 1377} & 0.83\,{\tiny 851} & 0.75\,{\tiny 944} & 1.09\,{\tiny 651} & 2.88\,{\tiny 246} & 1.46$\times$ \\
 & $8192 + 224{\times}256$ & 0.98\,{\tiny 852} & \textbf{0.97}\,{\tiny 860} & 2.12\,{\tiny 395} & 1.96\,{\tiny 427} & 3.11\,{\tiny 269} & 4.92\,{\tiny 170} & 2.01$\times$ \\
 & log-unif.\ 16k & 0.41\,{\tiny 1096} & \textbf{0.40}\,{\tiny 1132} & 0.57\,{\tiny 790} & 0.57\,{\tiny 788} & 0.60\,{\tiny 749} & 2.48\,{\tiny 182} & 1.43$\times$ \\
 & log-unif.\ 16k, GQA & 0.47\,{\tiny 968} & \textbf{0.46}\,{\tiny 986} & 0.70\,{\tiny 640} & 0.60\,{\tiny 749} & 0.64\,{\tiny 708} & {\tiny\texttt{no GQA}} & 1.32$\times$ \\
 & log-unif.\ 64k & 2.21\,{\tiny 1241} & \textbf{2.13}\,{\tiny 1281} & 2.79\,{\tiny 980} & 3.14\,{\tiny 872} & 3.39\,{\tiny 808} & 9.48\,{\tiny 289} & 1.31$\times$ \\
\bottomrule
\end{tabular}
\end{adjustbox}
\end{table}

\subsection{Performance on Packed Documents}
\label{sec:eval_varlen}

Each workload packs documents into one token row with 16 query heads and head dimension $d=128$. MHA uses 16 key/value heads; GQA 16/4 uses four. Equal-length workloads contain $4\times4096$, $16\times4096$, $16\times1024$, or $64\times1024$ tokens. Skewed workloads combine one 8192-token document with 32 or 224 documents of 256 tokens. Log-uniform workloads draw document lengths between 128 and 8192: 11 documents in 16k tokens, with and without GQA, and 29 documents in 64k tokens. We evaluate causal and non-causal attention; non-causal results appear in Appendix~\labelcref{app:noncausal}.

Attention training-call timing includes forward, backward, and quantization on packed inputs. Kernel-only measurements exclude quantization and use 128-aligned document starts for FLAMIA. Their layout difference prevents isolating quantization overhead by subtraction. For cuDNN FP8, input-scale computation is timed, whereas output scales are calibrated beforehand on the same data. Speedup denotes the fastest supported baseline's latency divided by FLAMIA's latency for the same workload.

MXFP8 achieves the lowest measured attention training-call latency on all nine causal workloads on both GPUs (\Cref{tab:varlen_train}), with speedups of $1.10$--$1.40\times$ on B200 and $1.06$--$1.42\times$ on B300. On non-causal workloads, MXFP8 leads on all nine B200 cases ($1.04$--$1.30\times$). B300 has five clear wins, two ties at displayed precision, and two small losses. Our non-causal backward leads on every workload, with speedups of $1.17$--$1.49\times$ and $1.01$--$1.43\times$ on B200 and B300, respectively; our forward leads on six and five of nine.

Kernel-only results show complementary strengths (\Cref{tab:varlen_fwd,tab:varlen_bwd} and Appendix~\labelcref{app:b300}). NVFP4 achieves the lowest causal backward latency on every workload, with speedups of $1.22$--$2.01\times$ on B200 and $1.17$--$2.00\times$ on B300. MXFP8 leads the forward comparisons on all nine B200 workloads and eight of nine B300 workloads, with ranges of $1.07$--$1.72\times$ and $0.98$--$1.69\times$. The NVFP4 forward is slower than MXFP8, consistent with its additional softmax work outweighing savings in $\mathbf{Q}\mathbf{K}^{\top}$.

B300/B200 training-call latency ratios are 0.85--0.93 for cuDNN FP8, 0.92--0.95 for FA4 BF16, and 0.89--0.94 for MXFP8. Relative advantages vary by workload rather than narrowing uniformly. The largest gains occur on $8192+224\times256$: B300 training-call and backward speedups reach $1.42\times$ and $2.00\times$. On this workload, cuDNN FP8 takes approximately $1.68\times$ our MXFP8 training-call latency on both GPUs, FA4 BF16 takes $1.40$--$1.42\times$, and cuDNN BF16 takes $3.7$--$3.9\times$. Equal-length workloads yield $1.06$--$1.18\times$ speedups. Skew (maximum/mean document length) is not isolated from other shape differences.

Multiplying one document's activations by $10^4$ gives a $\mathbf{dK}$ cosine of 0.99805 for the remaining documents with local scales, versus 0.99151 with tensor-wide scaling. At $T=131072$, measured peak memory per attention call is 2.7\,GB for packed inputs versus 16.8\,GB with buffers sized for aligned, padded documents. In full-model training of the 1B model on one B200, packed inputs achieve 84.6k and 96.8k tokens/s at packed row lengths of 16k and 64k, versus 68.8k and 84.3k for padding. The padded configuration exceeds available memory at 128k tokens.

\subsection{Numerical Accuracy}
\label{sec:eval_accuracy}

\begin{table}[!htbp]
\centering
\caption{Cosine to an FP32 reference on the same inputs, $\mathbf{O}$ / $\mathbf{dQ}$, $4 \times 4096$ tokens, 16 heads,
$d = 128$ (B200). \texttt{no bwd}: the system has no backward; \texttt{no causal mask}: the system supports only non-causal attention.}
\label{tab:accuracy}
\footnotesize
\setlength{\tabcolsep}{2pt}
\begin{tabular*}{\columnwidth}{@{\extracolsep{\fill}}lcc@{}}
\toprule
\textbf{System} & \textbf{causal} & \textbf{non-causal} \\
\midrule
Ours, MXFP8 (token-block) & 0.9988 / 0.9979 & 0.9985 / 0.9978 \\
Ours, NVFP4               & 0.9920 / 0.9906 & 0.9899 / 0.9890 \\
Ours, all-4-bit           & 0.9810 / 0.9456 & 0.9800 / 0.9509 \\
cuDNN FP8                 & 0.9985 / 0.9975 & 0.9982 / 0.9974 \\
cuDNN MXFP8 (1$\times$32) & 0.9988 / 0.9978 & 0.9985 / 0.9978 \\
LP-FA4 (1$\times$32)  & {\tiny\texttt{no causal mask}} & 0.9986 / 0.9978 \\
FlashInfer NVFP4 KV cache & 0.9909 / {\tiny\texttt{no bwd}} & 0.9895 / {\tiny\texttt{no bwd}} \\
HW-FP4 FA4, Direct-P &  {\tiny\texttt{no causal mask}} & 0.9435 / {\tiny\texttt{no bwd}} \\
\bottomrule
\end{tabular*}
\end{table}

\Cref{tab:accuracy} compares output and gradient cosine similarity with an FP32 reference on the same BF16 inputs. The evaluated 8-bit configurations achieve output cosines of approximately 0.998--0.999 across tensor-wide, per-32-element, and token-block scaling. On non-causal inputs, mixed NVFP4/FP8 achieves 0.9899, compared with 0.9895 for FlashInfer's NVFP4 KV-cache path and 0.9435 for the Direct-P forward of HW-FP4 FA4, which uses an MXFP4 $\mathbf{P}\mathbf{V}$. Different operands and scaling schemes confound bit-width comparisons. Within FLAMIA, all-4-bit $\mathbf{O}/\mathbf{dQ}$ cosines are 0.9810/0.9456 for causal attention and 0.9800/0.9509 for non-causal attention, versus 0.9920/0.9906 and 0.9899/0.9890 for mixed NVFP4/FP8. These metrics do not show training stability.

\subsection{Ablation Study}
\label{sec:eval_ablation}

Ablations disable individual optimizations on B200 unless stated otherwise. Gains are not additive.

Fused quantization reduces preprocessing from 789 to 261\,$\mu$s for MXFP8 and from 0.64 to 0.11\,ms for NVFP4 at $4\times4096$ tokens. The reported kernel launch counts decrease from 17 to one and from 11 to one, respectively. FLAMIA requires no separate output-scale calibration pass because outputs and returned gradients remain BF16. cuDNN FP8 calibration costs depend on which scales are computed online; Appendix~\labelcref{app:batched} compares three policies, including an additional forward/backward pass to obtain all scales.

A diagnostic forward with exponential evaluation removed reaches 2608\,TFLOP/s. The associated 45\,\% estimate is diagnostic: removal changes overlap and invalidates attention, precluding direct runtime attribution. Log-domain probability quantization yields reported NVFP4 forward gains of 14--19\,\%; splitting $\mathbf{Q}\mathbf{K}^{\top}$ yields 4.7\,\%, and folding score scales yields 4\,\% for MXFP8. Without the polynomial clamp, output cosine on a trained-model activation capture falls to $-0.012$. On B300, disabling exponential emulation in the pure-E4M3 configuration yields 17--33\,\% gains; fused \texttt{ld.red} on fully unmasked tiles yields 18--20\,\% for the causal NVFP4 forward.

The mixed NVFP4/FP8 backward reaches a geometric-mean throughput of 1329\,TFLOP/s, compared with 571 for all-4-bit and 1015 for BF16. Quantizing six copies accounts for 45--52\,\% of the all-4-bit backward time; its validation loss is 0.012 above BF16 in the reported training experiment. FP16 $\mathbf{dQ}$ partials with data-dependent scaling yield a reported 23.8\,\% backward gain. Without them, bulk reduce-add bandwidth near 6\,TB/s would limit the one-CTA backward to approximately 1990\,TFLOP/s. Checkerboard exponential assignment yields 6--10\,\% gains, and two-CTA execution yields 2.3--6.4\,\% at 32k keys or more. The optional causal $\mathbf{dQ}$ scale removes the sequence-wide accumulator dependency, not the remaining within-block dependencies. Its reported overhead is 8--11\,\% for backward and 6--9\,\% for the attention training call at 32768 tokens, largely from loading row scales during reduction.

Removing longest-first ordering incurs a reported 10--65\,\% causal-forward penalty. Merge-sort work-list construction takes 0.26\,ms for 131072 one-token documents, compared with 50.5\,ms for the reference construction. Just-in-time CLC acquisition improves performance by 1.7--5.7\,\% over static assignment in 13 of 16 evaluated causal configurations; premature acquisition incurs 6--14.5\,\% overhead. On non-causal workloads, CLC yields 4.1--9.7\,\% gains on the two skewed cases but loses on equal-length cases, motivating its activation when maximum document length is at least four times the mean. Removing document-aware backward ordering incurs up to 9.7\,\% overhead; removing head grouping for L2 locality incurs up to 22.5\,\%.

\subsection{Bit-Level Exponentials at Inference}
\label{sec:eval_inference}

We evaluate inference calls including quantization on B200 with 128-aligned document starts (Appendix~\labelcref{app:infexp}). Schraudolph's approximation gives MXFP8 speedups of approximately $1.03$--$1.17\times$, changing output cosine by less than $10^{-4}$. Mitchell-P matches exact-rounding codes on 98.7\,\% of evaluated inputs, versus 63.3\,\% for the affine classifier of Direct-P~\cite{hu2026hwfp4}. The complete Mitchell-P configuration achieves $1.05$--$1.23\times$ speedups over NVFP4 on non-causal workloads. On causal workloads, latency is approximately $1.25$--$1.77\times$ that of NVFP4, equivalent to 20--43\,\% lower throughput; diagonal per-key $\mathbf{V}$ scaling contributes to this overhead. Output cosine decreases from 0.990--0.993 to 0.980--0.982. Because Mitchell-P also changes probability precision, this comparison does not isolate the approximation.

The largest observed within-row score range is 39.6 nats on Gaussian inputs, within the approximation window, versus 386--1653 nats on trained-model captures, motivating the clamp. A Mitchell-P backward that consistently reconstructs the forward probabilities gives an attention training-call latency of 1.60--1.64\,ms at $4\times4096$ tokens, versus 1.00\,ms for the default NVFP4 path. In this configuration, backward overhead outweighs the forward benefit.

\subsection{Causality and Training Stability}
\label{sec:eval_numerics}

\begin{table}[t]
\centering
\caption{Largest relative change of earlier outputs when a future token raises the maximum of its scale group, on activations of a trained model, 8-bit / 4-bit route. The change spreads only within the group that shares the scale: the whole tensor, a 32-token block, or a group of 16 or 32 keys. \texttt{below noise}: smaller than the run-to-run variation of the 4-bit backward, which accumulates $\mathbf{dQ}$ with atomics and rounds stochastically.}
\label{tab:leaks}
\scriptsize
\setlength{\tabcolsep}{2pt}
\renewcommand{\arraystretch}{0.95}
\begin{tabular*}{\columnwidth}{@{\extracolsep{\fill}}lll@{}}
\toprule
\textbf{Scaling} & $\mathbf{O}$ & $\mathbf{dQ}$ \\
\midrule
Per-tensor & $9.8\mathrm{e}{-2}$ / $7.9\mathrm{e}{-2}$ & $2.8\mathrm{e}{-1}$ / $1.1\mathrm{e}{-1}$ \\
4-bit $\mathbf{V}$ by keys & $2.2\mathrm{e}{-1}$ & \texttt{below noise} \\
cuDNN MXFP8, default & $6.9\mathrm{e}{-6}$ & $1.1\mathrm{e}{-5}$ \\
cuDNN MXFP8, floor exp. & $1.1\mathrm{e}{-1}$ & $1.1\mathrm{e}{-1}$ \\
LP-FA4, default & $6.8\mathrm{e}{-6}$ & $8.5\mathrm{e}{-4}$ \\
LP-FA4, floor exp. & $1.0\mathrm{e}{-1}$ & $8.3\mathrm{e}{-2}$ \\
HW-FP4 FA4 (causal) & $1.7\mathrm{e}{-1}$ & {\tiny\texttt{no bwd}} \\
FlashInfer NVFP4 KV & $2.6\mathrm{e}{-1}$ & {\tiny\texttt{no bwd}} \\
\midrule
Ours & 0 / 0 & 0 / 0 \\
\bottomrule
\end{tabular*}
\end{table}

We probe scale-induced dependencies by perturbing a future key-position input to increase its scale-group maximum and measuring changes at earlier positions, on activations captured from a trained model. \Cref{tab:leaks} reports maximum relative changes in the evaluated probes. Leaks occur with per-tensor scales, with 4-bit $\mathbf{V}$ grouped along keys, in the causal path of HW-FP4 FA4, and with floor-rounded microscaling exponents. Upward rounding in cuDNN MXFP8 and Low Precision FA4 avoids the saturation mechanism, but both still change earlier outputs slightly. We measure the quantizers of Low Precision FA4 around exact attention, since its backward refuses causal masks.

Of all rows in \Cref{tab:leaks}, only our exact options with the causal $\mathbf{dQ}$ scale leave earlier outputs unchanged. The causal scale has a reported 8--11\,\% backward overhead. Applying per-key $\mathbf{V}$ scales only on diagonal tiles gives forward latency factors of $1.17\times$ and $1.04\times$ at 4096 and 32768 tokens, compared with $2.33\times$ and $2.83\times$ when applied to every tile. The $\mathbf{K}$ fold costs 2.7--4.0\,\% of the attention training call.

We train byte-level GPT models with six layers and width 1024 on \texttt{enwik8}~\cite{mahoney2011enwik8}, using 1024-byte contexts for 6000 steps. Models trained with the leaking 4-bit configuration achieve validation losses of 0.811--0.827 through their own kernels, compared with 0.852 for BF16. Evaluating the same weights with exact attention increases loss to 0.888--0.940, compared with 0.862 for a model trained without the leak.

Forward--backward mismatches destabilize the evaluated configurations. Using different $\mathbf{dO}$ representations for $D$ and $\mathbf{dP}$ causes divergence within 500 steps. Using 4-bit probabilities in forward and 8-bit probabilities in backward produces highly concentrated attention. Diagonal handling over 256 queries in forward versus 128 in backward increases gradient norm from 0.5 to $2\cdot10^5$, despite a single-step $\mathbf{dQ}$ cosine of 0.945. Approximating exponentials only in forward causes divergence by step 500. These failures motivate consistent numerical conventions.

In the $\mathbf{dS}$ scaling ablation, the fraction of entries rounded to zero rises from 0.4\,\% to 29\,\%, while the validation-loss gap to BF16 rises from 0.0008 to 0.049 despite similar reported norm error. This association supports preserving small gradient entries but does not isolate underflow as the sole cause. Deriving scales from each $32\times32$ block's maximum absolute value has a reported 9--12\,\% attention training-call overhead. Appendix~\labelcref{app:design} reports scale-range, rounding, and Hadamard-rotation effects.

\subsection{Training validation}
\label{sec:training_validation}

We pretrain 1B-parameter Llama-3-style Transformers~\cite{vaswani2017attention,grattafiori2024llama} on 100B FineWeb-Edu tokens~\cite{penedo2024fineweb} with next-token prediction and 2k-token contexts. Our TorchTitan integration~\cite{liang2025torchtitan} runs on eight B300 GPUs, and Appendix~\labelcref{app:training_details} lists the hyperparameters. We compare BF16 FlashAttention-4~\cite{zadouri2026fa4} with our MXFP8 and mixed NVFP4/FP8 paths. \Cref{tab:lm_eval_comparison} reports mean loss over the final 1,000 steps and final-checkpoint scores on five tasks using \texttt{lm-eval-harness} \cite{eval-harness}.

\begin{table}[!htb]
\centering
\caption{Training loss averaged over the final 1,000 steps and downstream evaluation of the final checkpoints.}
\label{tab:lm_eval_comparison}\label{tab:loss_comparison}
\footnotesize
\setlength{\tabcolsep}{4pt}
\renewcommand{\arraystretch}{0.9}
\begin{tabular*}{\columnwidth}{@{\extracolsep{\fill}}lrrr@{}}
\toprule
\textbf{Task} & \textbf{BF16} & \textbf{MXFP8} & \textbf{NVFP4} \\
 & Baseline & Ours & Ours \\
\midrule
Training loss $\downarrow$ & 2.392 & 2.397 & 2.406 \\
\midrule
ARC Challenge $\uparrow$ & 0.3925 & 0.3618 & 0.3677 \\
ARC Easy $\uparrow$     & 0.6481 & 0.6460 & 0.6557 \\
HellaSwag $\uparrow$    & 0.5711 & 0.5650 & 0.5610 \\
PIQA $\uparrow$         & 0.7280 & 0.7323 & 0.7258 \\
WinoGrande $\uparrow$   & 0.5762 & 0.5635 & 0.5967 \\
\midrule
\textbf{Mean} & \textbf{0.5832} & \textbf{0.5737} & \textbf{0.5814} \\
\bottomrule
\end{tabular*}
\end{table}

MXFP8 and NVFP4 quantization leads to marginal loss increase. Downstream scores for the low-bit attention are close to the baseline.

\section{Discussion and Conclusion}

We presented FLAMIA, a family of MXFP8 and mixed NVFP4/FP8 attention
kernels for NVIDIA Blackwell. FLAMIA combines token-block scaling,
forward--backward numerical consistency, and workload-aware scheduling
to process packed documents without repacking. Its MXFP8 path achieves
the lowest measured combined forward and backward attention latency,
including quantization, across the evaluated causal packed-document
workloads. 

Our experiments identify scale-induced dependencies, numerical
inconsistencies, and saturation that single-step similarity metrics
alone do not adequately characterize. Default scaling retains residual
dependencies; options that eliminate the identified dependencies
currently cover supported batched causal configurations. These findings
highlight the importance of jointly designing quantization and kernel
execution, and of evaluating low-precision attention through latency,
causal-dependency tests, and training outcomes.

\bibliographystyle{unsrtnat}
\bibliography{main}
\section*{AI Usage Disclosure}

We used the following generative AI tools in this work: Claude (Anthropic) through the Claude Code assistant, Codex (OpenAI), and the AI assistant built into Overleaf, where the manuscript was prepared for submission. These tools assisted us in writing attention kernels and polishing the text of the paper. The authors designed the methods, specified and reviewed every code change, and validated the code through tests and the measurements reported here. All research ideas, the design of the methods and experiments, and the conclusions are the authors' own. The authors take full responsibility for the content of this paper, including every part produced with AI assistance.
\appendix
\raggedbottom
\section{Design Details}
\label{app:design}

\subsection{Scale Ranges}

Attention's intermediates fall off the bottom of the scale formats. A per-16 UE4M3 scale for $\mathbf{P}$, the group maximum over 6, is zero once the group maximum drops below the smallest UE4M3 subnormal, which 63--89\,\% of the groups of a trained model do. Gradients have the same problem: an unscaled NVFP4 $\mathbf{dO}$ flushes every code at real gradient magnitudes, and \citet{hu2026hwfp4} report that an unscaled E4M3 $\mathbf{dO}$ rounds 97\,\% of values to zero. We therefore keep $\hat{\mathbf{P}}$ in 8 bits and give $\mathbf{dO}$ token-block scales.

\subsection{Hadamard Rotation}

One outlier channel sets the scale of every block it touches. A Hadamard rotation, the incoherence processing of QuIP~\cite{chee2024quip} and QuIP\#~\cite{Tseng2024QuIPEB}, of $\mathbf{Q}$ and $\mathbf{K}$ along the head dimension leaves $\mathbf{S}$ unchanged and spreads the outlier's energy over all channels. That helps a coarse shared scale and hurts a fine one: NVFP4's per-16 scale already isolates the outlier, and rotation spreads it into every group. The rotation adds no causal channel, since it mixes channels and never positions. With outlier channels in $\mathbf{K}$, it lowers the NVFP4 output cosine from 0.937 to 0.857 and raises the MXFP4 one from 0.749 to 0.816. In training, its effect stays within run-to-run noise on both routes. We fuse it into the quantizer, which reads each tensor once, and the forward and the backward read the same rotated codes. It is on by default wherever the head dimension is 128 and off on the all-4-bit route. Because training does not distinguish the two settings, we keep a single default rather than one per format.

\subsection{The In-Block Residue}

Token blocks and $32 \times 32$ $\mathbf{dS}$ blocks leave a small leak inside each block: a later token in the same block that raises its maximum re-rounds earlier subnormal codes, within at most 31 positions. We provide per-token operand scales and a per-query-row $\mathbf{dS}$ exponent, which close it exactly, as options rather than defaults: they cost backward time and did not change training. Together with the causal $\mathbf{dQ}$ scale described next, these options form the exact options of \Cref{sec:eval_numerics}.

\subsection{The \texorpdfstring{$\mathbf{dQ}$}{dQ} Accumulator}

We reduce the $\mathbf{dQ}$ partials in FP16, as the kernel of \citet{metalpfa4} does, which halves the reduction traffic but leaves only five exponent bits. A fixed boost overflows on outlier $\mathbf{K}$ channels, so we pre-scale the partials by a power of two sized from the maxima of $\|\mathbf{dO}\|$, $\|\mathbf{V}\|$ and $|\mathbf{K}|$. By default these maxima are taken over the whole sequence. A later key that raises one of them then changes the FP16 rounding of every earlier row of $\mathbf{dQ}$, by up to $5\times10^{-4}$ of its largest entry. The causal $\mathbf{dQ}$ scale removes this leak. It sizes the scale of query row $i$ from $\|\mathbf{dO}_i\|$ and from the running maxima of $\|\mathbf{V}_j\|$ and $|\mathbf{K}_j|$ over the keys $j \le i$, and every backward kernel reads the scale of its own row. It needs no host synchronization and keeps the call capturable in a CUDA graph. It makes the backward 8--11\,\% slower at 32768 tokens, so we provide it as an option. It currently covers batched causal calls. Packed documents, windows and the all-4-bit backward use the default scale.

\subsection{Rounding}

We round every element to nearest, with one exception. Stochastic rounding is unbiased, but it adds noise that the exponential amplifies on the softmax input, and it pays off only where nearest rounding flushes a gradient tail. Because block-scaled $\mathbf{dS}$ keeps that tail, we do not use it on the default routes. On the all-4-bit route we apply it to the $\mathbf{dK}$ operand with the hardware's stochastic convert, which lowers that route's validation loss from 0.943 to 0.921.

\section{Kernel Details}
\label{app:kernels}

\subsection{Quantizing \texorpdfstring{$\hat{\mathbf{P}}$}{P}}

Most probabilities are small, so we multiply $\mathbf{P}$ by a fixed factor $2^{c}$ before rounding to E4M3, which keeps small entries above the smallest E4M3 code, and divide the factor out after the MMA. FlashAttention-4 also uses conditional rescaling: it updates the running row maximum $m$ only when $m_j - m_{j-1} > \tau$, with $\tau = 8$ by default, so until the next update the exponentials can exceed 1 by up to $2^{\tau}$. In BF16 this is harmless. In E4M3, whose largest value is 448, $c + \tau$ must stay within $\log_2 448$. Otherwise the largest entries of $\hat{\mathbf{P}}$ saturate without any error, and $\mathbf{O}$ is wrong without a NaN. On the MXFP8 route we use $c = 4$ and $\tau = 4$, as cuDNN does, and assert the bound. This $\hat{\mathbf{P}}$ needs no other scale. A violation is easy to miss: with $c = 8$ and $\tau = 2$, entries of $\hat{\mathbf{P}}$ reach 1024, saturate at 448, and give a wrong $\mathbf{O}$ without a NaN. The bound costs no accuracy in training: with $(c, \tau) = (4, 4)$, a 6000-step run reaches a validation loss of 0.8493, and over four seeds an exact running maximum, $(c, \tau) = (8, 0)$, reaches 0.8567 against 0.8504 for BF16. In the backward, both routes multiply the $\hat{\mathbf{P}}$ of $\mathbf{dV}$ by $2^{8}$ and store no scale.

\subsection{4-Bit \texorpdfstring{$\mathbf{V}^{\top}$}{V Transpose} on Packed Rows}

On the all-4-bit route, $\mathbf{V}^{\top}$ stores two keys per byte. A document that starts at an odd key therefore starts in the middle of a byte, and so do its tiles. TMA and the MMA address memory in bytes, so such a tile cannot be loaded as is. Shifting every byte by half a byte in registers would combine two input bytes into each output byte on every tile. Instead, the quantizer writes $\mathbf{V}^{\top}$ twice, once with key pairs $(2m, 2m{+}1)$ and once with pairs $(2m{-}1, 2m)$. A document that starts at an odd key reads the second copy, where its tiles start on a byte boundary. Each tile is then loaded from the 16-byte boundary below it, as TMA requires, and moved into place in registers by whole bytes. The cost is one extra write of $\mathbf{V}^{\top}$ per step instead of a half-byte shuffle on every tile.

\subsection{Choosing the 2-CTA Backward}

We choose between the 1-CTA and 2-CTA backward with a rule fitted to measurements, and we refit it for each operand format, because a rule tuned for one format did not fit the next. From 32k keys on, the rule selects the 2-CTA kernel for every call it supports. Below 32k keys, it selects the 2-CTA kernel only for MHA with equal query and key lengths, where it measured faster. Deterministic calls keep the 1-CTA kernel.

\subsection{Inherited Mask}

Our FP8 forward had inherited the causal mask in its unmasked loop. Removing it speeds up causal forwards by 11--14\,\%.

\subsection{Determinism}

Our forward is deterministic: the same inputs give bitwise-identical $\mathbf{O}$ and LSE. By default, we accumulate $\mathbf{dQ}$ with atomics, as every backward we measured does, including BF16. A deterministic order costs 5--12\,\% for MHA and 20--34\,\% for GQA.

\subsection{Work Lists and Launch Layout}

We sort the forward's work list on the GPU with a fixed number of ordinary merge-sort launches, so it needs no host synchronization and no cooperative launch. In the backward we launch one CTA per position on a capped grid, sized from the row length, and a persistent overflow launch takes any later positions. In our tests, a single persistent kernel kept descriptors live across its loop and spilled registers.

\section{Inference-Only Exponentials}
\label{app:infexp}

\subsection{Mitchell-P}

One FFMA computes $t = (x - \delta)\,2^{16} + M$ with the magic constant $M = 2^{23} + L \cdot 2^{16}$. Adding $M$ fixes the exponent of $t$ at $2^{23}$, so the FFMA's rounding writes $x - \delta$ as a fixed-point number with 16 fractional bits into the low mantissa bits of $t$. The term $L \cdot 2^{16}$ offsets it so that negative scores stay in range. One integer shift-add, $\mathrm{bits}(\widetilde{2^{x}}) = (\mathrm{bits}(t) \ll 7) + c$ with $c = (127 \ll 23) - (\mathrm{bits}(M) \ll 7)$, moves the integer part of $x$ from bit 16 into the exponent field and its fraction into the mantissa, which yields the piecewise-linear approximation $2^{\lfloor x \rfloor}(1 + \{x\})$. This approximation never falls below $2^{x}$, and the offset $\delta = \log_2 1.25 - 0.25$ shifts it down to centre its error. The FFMA is exact while $x - \delta \in [-L, 128 - L)$. With $L = 124$, scores down to about 86 nats below their group maximum are exact. Below this window the shifted pattern would wrap, so one integer maximum clamps $\mathrm{bits}(t)$ at $\mathrm{bits}(2^{23})$, which makes the result exact for any range. Within the window the value differs from $2^{x}$ by at most 4.7\,\%, and we accumulate the row sum from the same values, so the backward of the all-4-bit route can rebuild them exactly.

\subsection{Schraudolph's Exponential}

The FFMA computes $t = 2^{7}x + 1.5 \cdot 2^{23} + 2^{7}(127 - \sigma)$ with $\sigma = 0.0573$, and its low 16 bits are shifted left by 16 bits into a float and clamped at zero. The magic constant $1.5 \cdot 2^{23}$ rounds $2^{7}x$ to an integer in the low mantissa bits of $t$, and the low 16 bits of $t$ read as BF16 give $2^{\lfloor x \rfloor}(1 + \{x\})$ shifted by $\sigma$. Its ratio to $2^{x}$ stays between $-3.9\,\%$ and $+2.0\,\%$ and averages one, and the row sum and the log-sum-exp therefore carry no mean bias. The 16-bit field wraps below $x = -383$, far outside any score range we observed.

\begin{table}[!htbp]
\centering
\caption{Inference call on packed documents with the bit-level exponentials (B200, 16 heads, $d = 128$): time in ms. Schraudolph: our MXFP8 route with Schraudolph's exponential, change against MXFP8 in small type; Mitchell-P: the 4-bit $\hat{\mathbf{P}}$ route with Mitchell-P, change against our NVFP4 route. Output cosine: 0.9985--0.9990 for MXFP8 with or without Schraudolph's exponential, 0.980--0.982 for Direct-P against 0.990--0.993 for NVFP4.}\label{tab:infexp}
\scriptsize
\setlength{\tabcolsep}{3pt}
\renewcommand{\arraystretch}{1.0}
\begin{adjustbox}{max width=\linewidth,max totalheight=0.78\textheight}
\begin{tabular}{@{}lllllllll@{}}
\toprule
& \multicolumn{4}{l}{\textbf{Causal}} & \multicolumn{4}{l}{\textbf{Non-causal}} \\
\cmidrule(lr){2-5}\cmidrule(lr){6-9}
\textbf{Documents} & MXFP8 & Schraudolph & NVFP4 & Mitchell-P & MXFP8 & Schraudolph & NVFP4 & Mitchell-P \\
\midrule
$16{\times}1024$ & 0.159 & 0.154\,{\tiny +3\,\%} & 0.182 & 0.312\,{\tiny $-$42\,\%} & 0.183 & 0.169\,{\tiny +8\,\%} & 0.221 & 0.210\,{\tiny +5\,\%} \\
$64{\times}1024$ & 0.527 & 0.514\,{\tiny +3\,\%} & 0.627 & 1.109\,{\tiny $-$43\,\%} & 0.640 & 0.591\,{\tiny +8\,\%} & 0.793 & 0.737\,{\tiny +8\,\%} \\
$4{\times}4096$ & 0.261 & 0.244\,{\tiny +7\,\%} & 0.318 & 0.419\,{\tiny $-$24\,\%} & 0.399 & 0.345\,{\tiny +16\,\%} & 0.507 & 0.419\,{\tiny +21\,\%} \\
$16{\times}4096$ & 1.033 & 0.943\,{\tiny +10\,\%} & 1.178 & 1.542\,{\tiny $-$24\,\%} & 1.632 & 1.429\,{\tiny +14\,\%} & 1.935 & 1.574\,{\tiny +23\,\%} \\
$8192 + 32{\times}256$ & 0.269 & 0.248\,{\tiny +8\,\%} & 0.329 & 0.429\,{\tiny $-$23\,\%} & 0.448 & 0.383\,{\tiny +17\,\%} & 0.570 & 0.467\,{\tiny +22\,\%} \\
$8192 + 224{\times}256$ & 0.567 & 0.549\,{\tiny +3\,\%} & 0.676 & 1.142\,{\tiny $-$41\,\%} & 0.745 & 0.682\,{\tiny +9\,\%} & 0.919 & 0.843\,{\tiny +9\,\%} \\
log-unif.\ 16k & 0.226 & 0.214\,{\tiny +6\,\%} & 0.272 & 0.401\,{\tiny $-$32\,\%} & 0.318 & 0.283\,{\tiny +12\,\%} & 0.398 & 0.342\,{\tiny +16\,\%} \\
log-unif.\ 16k, GQA & 0.193 & 0.181\,{\tiny +7\,\%} & 0.232 & 0.289\,{\tiny $-$20\,\%} & 0.285 & 0.250\,{\tiny +14\,\%} & 0.358 & 0.297\,{\tiny +21\,\%} \\
log-unif.\ 64k & 1.063 & 0.993\,{\tiny +7\,\%} & 1.224 & 1.626\,{\tiny $-$25\,\%} & 1.692 & 1.516\,{\tiny +12\,\%} & 1.983 & 1.629\,{\tiny +22\,\%} \\
\bottomrule
\end{tabular}
\end{adjustbox}
\end{table}

\section{Hardware and Software}
\label{app:hardware}

The kernels, the TorchTitan integration and the scripts that reproduce the benchmark tables are available at \url{\codeurl}. Our 4-bit forward kernel starts from the FP4 FlashAttention-4 forward released with Attn-QAT~\cite{attnqat} at \url{https://github.com/hao-ai-lab/flash-attention-fp4}, which provides its base structure. The optimizations on top of this base are our own: log-domain probability quantization, the split of the next tile's $\mathbf{Q}\mathbf{K}^{\top}$ around $\hat{\mathbf{P}}\mathbf{V}$ (Section~\labelcref{sec:kernels}) and the bit-level exponentials. So are the 8-bit route, the backward kernels, the packed-document support and the causal-safety mechanisms.

The kernel benchmarks run on NVIDIA B200 (\texttt{sm\_100a}) and B300 SXM6 (\texttt{sm\_103a}) GPUs, both with 148 SMs, at free-running clocks, with CUDA 13.0, PyTorch 2.14 and CuTe DSL 4.8. The two parts differ in the units our kernels lean on: B300 issues MUFU exponentials at 32 per clock per SM against 16 on B200. Both have 512 TMEM columns per SM and 232\,KB of shared memory per SM.

\subsection{Timing}

Each time is the median of three paired rounds after a discarded warm-up round. Within a round, the jobs of a row run back to back on the same GPU, in reversed order on every other round, and each call is captured in a CUDA graph. cuDNN FP8 takes all of its scales from the caller. In the training call we compute its $\mathbf{Q}$, $\mathbf{K}$, $\mathbf{V}$ and $\mathbf{dO}$ scales inside the timed call, as we do for our kernels and as a training step on new tensors must. Its output scales must exist before the call, and we calibrate them beforehand on the same data.

\section{Benchmarking Pitfalls}
\label{app:pitfalls}

Several effects silently invalidate low-precision attention benchmarks, and each one affected our own measurements before we controlled for it.
The benchmark harness of Low Precision FA4 applies no causal discount to FLOPs, so causal throughput in its convention is twice the true value. In its convention, our causal rows read 2617--4885 TFLOP/s, above the GPU's peak.

cuDNN's BF16 ragged backward returns wrong gradients when the softmax statistics are padded to the batch shape instead of ragged (dQ cosine 0.62 causal, 0.93 non-causal).

cuDNN's MXFP8 graph has one $\mathbf{V}$ input, but the forward needs $\mathbf{V}$'s blocks along keys and the backward along the head dimension. Feeding the forward's layout to the backward gives dQ and dK cosines of 0.947--0.962 while dV stays at 0.9997.

The same kernel source ran 18.6--20.6\,\% faster or slower under another CuTe DSL version with bitwise-identical output, so every timing is tied to its toolchain.

Replacing a plain TMEM load with SM103's fused-maximum load changes the instruction stream, and ptxas then contracts other operations differently. Outputs can then change in the last digits even where the loaded values are equal. We therefore gate every speed change bitwise per route, not per kernel family.

\section{System Integration}
\label{app:systems}

Both entry points run as \texttt{torch.library} custom operators, so selective activation checkpointing can save the attention outputs instead of re-running the forward and its quantizers in the backward, and \texttt{torch.compile} traces them through a fake implementation. No entry point synchronizes with the host: scales, work lists and stochastic-rounding seeds live on the GPU, so every call can be captured in a CUDA graph, and graph replay is bitwise equal to eager execution.

\section{Training Details}
\label{app:training_details}

We provide additional details of the training setup used in \Cref{sec:training_validation}. All runs use the hyperparameters listed in \Cref{tab:training_hyperparameters}. The training experiments use PyTorch~2.15.0dev20260928 with the CUDA~13.1 runtime.

\begin{table}[!htb]
\centering
\caption{Training hyperparameters shared by all attention variants.}
\label{tab:training_hyperparameters}
\small
\setlength{\tabcolsep}{4pt}
\renewcommand{\arraystretch}{1.0}
\begin{tabular*}{\columnwidth}{@{\extracolsep{\fill}}ll@{}}
\toprule
\textbf{Hyperparameter} & \textbf{Value} \\
\midrule
Global batch size & $2^{20}$ tokens \\
Training steps & 100,000 \\
\midrule
Optimizer & AdamW \\
Weight decay & 0.1 \\
\midrule
Peak learning rate & $3 \times 10^{-4}$ \\
Warmup steps & 300 \\
Learning-rate decay & Linear to zero \\
\bottomrule
\end{tabular*}
\end{table}

\section{Non-Causal Packed Documents}
\label{app:noncausal}

\Cref{tab:varlen_train_nc,tab:varlen_fwd_nc,tab:varlen_bwd_nc} repeat the measurements of \Cref{sec:eval_varlen} on the non-causal version of every workload, where Low Precision FA4 also runs.

\begin{table}[!htbp]
\centering
\caption{Training call on packed documents, non-causal, 16 heads, $d = 128$: time in ms (fastest in bold), TFLOP/s in small type; Speedup: the fastest baseline's time over ours.}\label{tab:varlen_train_nc}
\scriptsize
\setlength{\tabcolsep}{2pt}
\renewcommand{\arraystretch}{1.0}
\begin{adjustbox}{max width=\linewidth,max totalheight=0.78\textheight}
\begin{tabular}{@{}llllllllll@{}}
\toprule
& & \multicolumn{2}{l}{\textbf{Ours}} & \multicolumn{2}{l}{\textbf{FP8}} & \multicolumn{3}{l}{\textbf{BF16}} & \\
\cmidrule(lr){3-4}\cmidrule(lr){5-6}\cmidrule(lr){7-9}
\textbf{GPU} & \textbf{Documents} & MXFP8 & NVFP4 & cuDNN & LP-FA4 & FA4 & cuDNN & SDPA & \textbf{Speedup} \\
\midrule
B200 & $4{\times}4096$ & \textbf{1.46}\,{\tiny 1315} & 1.48\,{\tiny 1299} & 1.57\,{\tiny 1221} & 5.97\,{\tiny 322} & 1.76\,{\tiny 1096} & 1.98\,{\tiny 970} & 7.36\,{\tiny 261} & 1.08$\times$ \\
 & $16{\times}4096$ & \textbf{5.80}\,{\tiny 1326} & 5.86\,{\tiny 1312} & 6.17\,{\tiny 1248} & 23.3\,{\tiny 331} & 6.96\,{\tiny 1105} & 7.77\,{\tiny 991} & 23.5\,{\tiny 328} & 1.06$\times$ \\
 & $16{\times}1024$ & \textbf{0.56}\,{\tiny 864} & 0.58\,{\tiny 832} & 0.59\,{\tiny 809} & 6.82\,{\tiny 71} & 0.64\,{\tiny 753} & 0.65\,{\tiny 741} & 6.00\,{\tiny 80} & 1.07$\times$ \\
 & $64{\times}1024$ & \textbf{2.15}\,{\tiny 895} & 2.20\,{\tiny 874} & 2.24\,{\tiny 858} & 26.8\,{\tiny 72} & 2.59\,{\tiny 744} & 2.69\,{\tiny 715} & 7.71\,{\tiny 250} & 1.04$\times$ \\
 & $8192 + 32{\times}256$ & \textbf{1.54}\,{\tiny 1288} & 1.56\,{\tiny 1272} & 1.72\,{\tiny 1152} & 10.5\,{\tiny 189} & 1.80\,{\tiny 1099} & 2.18\,{\tiny 912} & 7.33\,{\tiny 271} & 1.12$\times$ \\
 & $8192 + 224{\times}256$ & \textbf{2.47}\,{\tiny 949} & 2.51\,{\tiny 933} & 3.49\,{\tiny 671} & 50.8\,{\tiny 46} & 3.22\,{\tiny 728} & 7.08\,{\tiny 331} & 10.2\,{\tiny 231} & 1.30$\times$ \\
 & log-unif.\ 16k & \textbf{1.11}\,{\tiny 1136} & 1.12\,{\tiny 1124} & 1.19\,{\tiny 1063} & 6.89\,{\tiny 183} & 1.33\,{\tiny 949} & 1.47\,{\tiny 859} & 6.17\,{\tiny 204} & 1.07$\times$ \\
 & log-unif.\ 16k, GQA & \textbf{1.07}\,{\tiny 1175} & 1.10\,{\tiny 1144} & 1.27\,{\tiny 989} & {\tiny\texttt{no GQA}} & 1.33\,{\tiny 945} & 1.50\,{\tiny 841} & {\tiny\texttt{no GQA}} & 1.19$\times$ \\
 & log-unif.\ 64k & \textbf{5.97}\,{\tiny 1283} & 6.03\,{\tiny 1270} & 6.38\,{\tiny 1200} & 26.2\,{\tiny 292} & 7.17\,{\tiny 1068} & 8.17\,{\tiny 937} & 24.1\,{\tiny 317} & 1.07$\times$ \\
\midrule
B300 & $4{\times}4096$ & \textbf{1.31}\,{\tiny 1468} & 1.33\,{\tiny 1442} & 1.34\,{\tiny 1438} & 5.61\,{\tiny 344} & 1.71\,{\tiny 1128} & 1.88\,{\tiny 1022} & 8.38\,{\tiny 238} & 1.02$\times$ \\
 & $16{\times}4096$ & \textbf{5.09}\,{\tiny 1512} & 5.19\,{\tiny 1483} & 5.09\,{\tiny 1511} & 21.9\,{\tiny 352} & 6.51\,{\tiny 1182} & 7.19\,{\tiny 1072} & 23.7\,{\tiny 327} & 1.00$\times$ \\
 & $16{\times}1024$ & \textbf{0.51}\,{\tiny 941} & 0.54\,{\tiny 893} & 0.52\,{\tiny 925} & 6.42\,{\tiny 75} & 0.60\,{\tiny 797} & 0.60\,{\tiny 804} & 9.04\,{\tiny 53} & 1.02$\times$ \\
 & $64{\times}1024$ & 1.95\,{\tiny 989} & 2.02\,{\tiny 954} & \textbf{1.94}\,{\tiny 991} & 25.5\,{\tiny 76} & 2.42\,{\tiny 861} & 2.48\,{\tiny 776} & 9.19\,{\tiny 209} & 1.00$\times$ \\
 & $8192 + 32{\times}256$ & \textbf{1.36}\,{\tiny 1461} & 1.40\,{\tiny 1416} & 1.44\,{\tiny 1373} & 9.64\,{\tiny 206} & 1.69\,{\tiny 1172} & 2.03\,{\tiny 983} & 9.43\,{\tiny 217} & 1.06$\times$ \\
 & $8192 + 224{\times}256$ & \textbf{2.25}\,{\tiny 1043} & 2.34\,{\tiny 1004} & 3.13\,{\tiny 748} & 47.7\,{\tiny 49} & 2.99\,{\tiny 784} & 6.82\,{\tiny 344} & 10.6\,{\tiny 220} & 1.33$\times$ \\
 & log-unif.\ 16k & \textbf{1.01}\,{\tiny 1253} & 1.02\,{\tiny 1234} & 1.01\,{\tiny 1248} & 6.50\,{\tiny 194} & 1.27\,{\tiny 989} & 1.39\,{\tiny 905} & 8.75\,{\tiny 144} & 1.00$\times$ \\
 & log-unif.\ 16k, GQA & \textbf{0.96}\,{\tiny 1308} & 1.02\,{\tiny 1239} & 1.08\,{\tiny 1170} & {\tiny\texttt{no GQA}} & 1.22\,{\tiny 1035} & 1.37\,{\tiny 922} & {\tiny\texttt{no GQA}} & 1.12$\times$ \\
 & log-unif.\ 64k & 5.25\,{\tiny 1458} & 5.35\,{\tiny 1431} & \textbf{5.24}\,{\tiny 1461} & 24.9\,{\tiny 307} & 6.74\,{\tiny 1137} & 7.59\,{\tiny 1009} & 24.3\,{\tiny 322} & 1.00$\times$ \\
\bottomrule
\end{tabular}
\end{adjustbox}
\end{table}

\begin{table}[!htbp]
\centering
\caption{Forward kernel on packed documents, non-causal, 16 heads, $d = 128$: time in ms (fastest in bold), TFLOP/s in small type; Speedup: the fastest baseline's time over ours. cuDNN MX: cuDNN MXFP8; LP-FA4: Low Precision FA4; FlashInfer: its fastest backend;}\label{tab:varlen_fwd_nc}
\scriptsize
\setlength{\tabcolsep}{1pt}
\renewcommand{\arraystretch}{1.0}
\begin{adjustbox}{max width=\linewidth,max totalheight=0.78\textheight}
\begin{tabular}{@{}lllllllllllllll@{}}
\toprule
& & \multicolumn{2}{l}{\textbf{Ours}} & \multicolumn{5}{l}{\textbf{8-bit}} & \textbf{4-bit} & \multicolumn{4}{l}{\textbf{BF16}} & \\
\cmidrule(lr){3-4}\cmidrule(lr){5-9}\cmidrule(lr){10-10}\cmidrule(lr){11-14}
\textbf{GPU} & \textbf{Documents} & MXFP8 & NVFP4 & \shortstack{cuDNN\\FP8} & \shortstack{cuDNN\\MX} & FA4 FP8 & FlashInfer & LP-FA4 & FlashInfer & FA4 & cuDNN & FlashInfer & SDPA & \textbf{Speedup} \\
\midrule
B200 & $4{\times}4096$ & \shortstack{0.31\\{\tiny 1795}} & \shortstack{0.40\\{\tiny 1373}} & \shortstack{0.31\\{\tiny 1757}} & \shortstack{0.57\\{\tiny 966}} & \shortstack{0.47\\{\tiny 1167}} & \shortstack{\textbf{0.29}\\{\tiny 1898}} & \shortstack{0.32\\{\tiny 1719}} & \shortstack{0.36\\{\tiny 1521}} & \shortstack{0.40\\{\tiny 1384}} & \shortstack{0.58\\{\tiny 953}} & \shortstack{0.40\\{\tiny 1373}} & \shortstack{2.37\\{\tiny 232}} & 0.95$\times$ \\
 & $16{\times}4096$ & \shortstack{1.32\\{\tiny 1666}} & \shortstack{1.58\\{\tiny 1392}} & \shortstack{1.31\\{\tiny 1681}} & \shortstack{2.24\\{\tiny 981}} & \shortstack{1.87\\{\tiny 1176}} & \shortstack{\textbf{1.21}\\{\tiny 1811}} & \shortstack{1.42\\{\tiny 1553}} & \shortstack{1.46\\{\tiny 1510}} & \shortstack{1.78\\{\tiny 1238}} & \shortstack{2.22\\{\tiny 989}} & \shortstack{1.74\\{\tiny 1261}} & \shortstack{6.51\\{\tiny 338}} & 0.92$\times$ \\
 & $16{\times}1024$ & \shortstack{\textbf{0.10}\\{\tiny 1452}} & \shortstack{0.12\\{\tiny 1191}} & \shortstack{0.11\\{\tiny 1244}} & \shortstack{0.25\\{\tiny 548}} & \shortstack{0.15\\{\tiny 911}} & \shortstack{0.11\\{\tiny 1302}} & \shortstack{0.11\\{\tiny 1275}} & \shortstack{0.12\\{\tiny 1176}} & \shortstack{0.13\\{\tiny 1044}} & \shortstack{0.19\\{\tiny 721}} & \shortstack{0.13\\{\tiny 1044}} & \shortstack{1.89\\{\tiny 73}} & 1.12$\times$ \\
 & $64{\times}1024$ & \shortstack{\textbf{0.35}\\{\tiny 1573}} & \shortstack{0.44\\{\tiny 1247}} & \shortstack{0.37\\{\tiny 1496}} & \shortstack{1.11\\{\tiny 494}} & \shortstack{0.60\\{\tiny 918}} & \shortstack{0.42\\{\tiny 1321}} & \shortstack{0.37\\{\tiny 1488}} & \shortstack{0.44\\{\tiny 1244}} & \shortstack{0.53\\{\tiny 1037}} & \shortstack{0.91\\{\tiny 605}} & \shortstack{0.52\\{\tiny 1052}} & \shortstack{2.40\\{\tiny 229}} & 1.05$\times$ \\
 & $8192 + 32{\times}256$ & \shortstack{\textbf{0.34}\\{\tiny 1665}} & \shortstack{0.45\\{\tiny 1274}} & \shortstack{0.37\\{\tiny 1516}} & \shortstack{0.60\\{\tiny 947}} & \shortstack{0.51\\{\tiny 1110}} & \shortstack{0.47\\{\tiny 1220}} & \shortstack{0.41\\{\tiny 1393}} & \shortstack{0.43\\{\tiny 1324}} & \shortstack{0.42\\{\tiny 1347}} & \shortstack{0.53\\{\tiny 1072}} & \shortstack{0.52\\{\tiny 1086}} & \shortstack{2.48\\{\tiny 229}} & 1.10$\times$ \\
 & $8192 + 224{\times}256$ & \shortstack{\textbf{0.42}\\{\tiny 1588}} & \shortstack{0.52\\{\tiny 1296}} & \shortstack{0.57\\{\tiny 1173}} & \shortstack{3.51\\{\tiny 191}} & \shortstack{0.79\\{\tiny 852}} & \shortstack{0.73\\{\tiny 922}} & \shortstack{0.48\\{\tiny 1384}} & \shortstack{0.93\\{\tiny 719}} & \shortstack{0.69\\{\tiny 970}} & \shortstack{3.48\\{\tiny 192}} & \shortstack{0.82\\{\tiny 820}} & \shortstack{3.12\\{\tiny 215}} & 1.15$\times$ \\
 & log-unif.\ 16k & \shortstack{\textbf{0.22}\\{\tiny 1606}} & \shortstack{0.29\\{\tiny 1244}} & \shortstack{0.25\\{\tiny 1462}} & \shortstack{0.46\\{\tiny 786}} & \shortstack{0.36\\{\tiny 999}} & \shortstack{0.25\\{\tiny 1427}} & \shortstack{0.28\\{\tiny 1285}} & \shortstack{0.30\\{\tiny 1203}} & \shortstack{0.30\\{\tiny 1201}} & \shortstack{0.42\\{\tiny 846}} & \shortstack{0.32\\{\tiny 1141}} & \shortstack{2.01\\{\tiny 179}} & 1.10$\times$ \\
 & log-unif.\ 16k, GQA & \shortstack{\textbf{0.22}\\{\tiny 1604}} & \shortstack{0.29\\{\tiny 1244}} & \shortstack{0.25\\{\tiny 1464}} & \shortstack{0.46\\{\tiny 788}} & \shortstack{0.28\\{\tiny 1288}} & \shortstack{0.35\\{\tiny 1034}} & {\tiny\texttt{no GQA}} & \shortstack{0.30\\{\tiny 1205}} & \shortstack{0.29\\{\tiny 1250}} & \shortstack{0.44\\{\tiny 813}} & \shortstack{0.38\\{\tiny 941}} & {\tiny\texttt{no GQA}} & 1.10$\times$ \\
 & log-unif.\ 64k & \shortstack{1.36\\{\tiny 1611}} & \shortstack{1.62\\{\tiny 1349}} & \shortstack{\textbf{1.35}\\{\tiny 1625}} & \shortstack{2.43\\{\tiny 899}} & \shortstack{1.93\\{\tiny 1137}} & \shortstack{1.44\\{\tiny 1518}} & \shortstack{1.44\\{\tiny 1520}} & \shortstack{1.55\\{\tiny 1414}} & \shortstack{1.82\\{\tiny 1203}} & \shortstack{2.40\\{\tiny 912}} & \shortstack{1.93\\{\tiny 1134}} & \shortstack{6.65\\{\tiny 329}} & 0.99$\times$ \\
\midrule
B300 & $4{\times}4096$ & \shortstack{0.24\\{\tiny 2317}} & \shortstack{0.32\\{\tiny 1733}} & \shortstack{0.23\\{\tiny 2376}} & \shortstack{0.23\\{\tiny 2371}} & \shortstack{0.31\\{\tiny 1769}} & \shortstack{\textbf{0.23}\\{\tiny 2398}} & \shortstack{0.30\\{\tiny 1832}} & \shortstack{0.32\\{\tiny 1711}} & \shortstack{0.36\\{\tiny 1513}} & \shortstack{0.54\\{\tiny 1010}} & \shortstack{0.36\\{\tiny 1514}} & \shortstack{3.13\\{\tiny 228}} & 0.97$\times$ \\
 & $16{\times}4096$ & \shortstack{0.92\\{\tiny 2358}} & \shortstack{1.25\\{\tiny 1762}} & \shortstack{\textbf{0.84}\\{\tiny 2429}} & \shortstack{0.95\\{\tiny 2326}} & \shortstack{1.23\\{\tiny 1795}} & \shortstack{0.95\\{\tiny 2320}} & \shortstack{1.17\\{\tiny 1904}} & \shortstack{1.27\\{\tiny 1734}} & \shortstack{1.56\\{\tiny 1408}} & \shortstack{2.05\\{\tiny 1071}} & \shortstack{1.57\\{\tiny 1398}} & \shortstack{6.96\\{\tiny 318}} & 0.97$\times$ \\
 & $16{\times}1024$ & \shortstack{\textbf{0.08}\\{\tiny 1785}} & \shortstack{0.10\\{\tiny 1448}} & \shortstack{0.09\\{\tiny 1533}} & \shortstack{0.08\\{\tiny 1715}} & \shortstack{0.11\\{\tiny 1244}} & \shortstack{0.09\\{\tiny 1585}} & \shortstack{0.09\\{\tiny 1496}} & \shortstack{0.10\\{\tiny 1330}} & \shortstack{0.12\\{\tiny 1101}} & \shortstack{0.18\\{\tiny 742}} & \shortstack{0.12\\{\tiny 1128}} & \shortstack{2.56\\{\tiny 46}} & 1.04$\times$ \\
 & $64{\times}1024$ & \shortstack{0.29\\{\tiny 1918}} & \shortstack{0.36\\{\tiny 1520}} & \shortstack{0.28\\{\tiny 1944}} & \shortstack{\textbf{0.28}\\{\tiny 1978}} & \shortstack{0.44\\{\tiny 1256}} & \shortstack{0.33\\{\tiny 1660}} & \shortstack{0.34\\{\tiny 1643}} & \shortstack{0.39\\{\tiny 1396}} & \shortstack{0.49\\{\tiny 1113}} & \shortstack{0.84\\{\tiny 651}} & \shortstack{0.47\\{\tiny 1182}} & \shortstack{2.92\\{\tiny 192}} & 0.97$\times$ \\
 & $8192 + 32{\times}256$ & \shortstack{\textbf{0.27}\\{\tiny 2112}} & \shortstack{0.36\\{\tiny 1568}} & \shortstack{0.27\\{\tiny 2096}} & \shortstack{0.41\\{\tiny 1385}} & \shortstack{0.33\\{\tiny 1736}} & \shortstack{0.39\\{\tiny 1445}} & \shortstack{0.33\\{\tiny 1705}} & \shortstack{0.38\\{\tiny 1490}} & \shortstack{0.39\\{\tiny 1463}} & \shortstack{0.53\\{\tiny 1076}} & \shortstack{0.44\\{\tiny 1297}} & \shortstack{2.87\\{\tiny 156}} & 1.01$\times$ \\
 & $8192 + 224{\times}256$ & \shortstack{\textbf{0.36}\\{\tiny 1860}} & \shortstack{0.46\\{\tiny 1475}} & \shortstack{0.47\\{\tiny 1440}} & \shortstack{1.53\\{\tiny 439}} & \shortstack{0.58\\{\tiny 1150}} & \shortstack{0.62\\{\tiny 1075}} & \shortstack{0.44\\{\tiny 1506}} & \shortstack{0.82\\{\tiny 821}} & \shortstack{0.64\\{\tiny 1046}} & \shortstack{3.37\\{\tiny 199}} & \shortstack{0.68\\{\tiny 981}} & \shortstack{3.37\\{\tiny 199}} & 1.24$\times$ \\
 & log-unif.\ 16k & \shortstack{\textbf{0.18}\\{\tiny 2020}} & \shortstack{0.23\\{\tiny 1560}} & \shortstack{0.18\\{\tiny 1949}} & \shortstack{0.21\\{\tiny 1754}} & \shortstack{0.24\\{\tiny 1498}} & \shortstack{0.20\\{\tiny 1769}} & \shortstack{0.26\\{\tiny 1394}} & \shortstack{0.27\\{\tiny 1347}} & \shortstack{0.28\\{\tiny 1290}} & \shortstack{0.43\\{\tiny 843}} & \shortstack{0.29\\{\tiny 1223}} & \shortstack{2.84\\{\tiny 126}} & 1.04$\times$ \\
 & log-unif.\ 16k, GQA & \shortstack{\textbf{0.17}\\{\tiny 2064}} & \shortstack{0.23\\{\tiny 1564}} & \shortstack{0.18\\{\tiny 1992}} & \shortstack{0.20\\{\tiny 1763}} & \shortstack{0.24\\{\tiny 1493}} & \shortstack{0.30\\{\tiny 1185}} & {\tiny\texttt{no GQA}} & \shortstack{0.27\\{\tiny 1355}} & \shortstack{0.25\\{\tiny 1450}} & \shortstack{0.41\\{\tiny 880}} & \shortstack{0.33\\{\tiny 1102}} & {\tiny\texttt{no GQA}} & 1.04$\times$ \\
 & log-unif.\ 64k & \shortstack{0.95\\{\tiny 2300}} & \shortstack{1.27\\{\tiny 1716}} & \shortstack{\textbf{0.93}\\{\tiny 2361}} & \shortstack{1.04\\{\tiny 2094}} & \shortstack{1.27\\{\tiny 1723}} & \shortstack{1.06\\{\tiny 2037}} & \shortstack{1.25\\{\tiny 1754}} & \shortstack{1.36\\{\tiny 1609}} & \shortstack{1.64\\{\tiny 1334}} & \shortstack{2.37\\{\tiny 923}} & \shortstack{1.74\\{\tiny 1283}} & \shortstack{7.12\\{\tiny 315}} & 0.97$\times$ \\
\bottomrule
\end{tabular}
\end{adjustbox}
\end{table}

\begin{table}[!htbp]
\centering
\caption{Backward kernels on packed documents, non-causal, 16 heads, $d = 128$: time in ms (fastest in bold), TFLOP/s in small type; Speedup: the fastest baseline's time over ours.}\label{tab:varlen_bwd_nc}
\scriptsize
\setlength{\tabcolsep}{2pt}
\renewcommand{\arraystretch}{1.0}
\begin{adjustbox}{max width=\linewidth,max totalheight=0.78\textheight}
\begin{tabular}{@{}llllllllll@{}}
\toprule
& & \multicolumn{2}{l}{\textbf{Ours}} & \multicolumn{2}{l}{\textbf{FP8}} & \multicolumn{3}{l}{\textbf{BF16}} & \\
\cmidrule(lr){3-4}\cmidrule(lr){5-6}\cmidrule(lr){7-9}
\textbf{GPU} & \textbf{Documents} & MXFP8 & NVFP4 & cuDNN & LP-FA4 & FA4 & cuDNN & SDPA & \textbf{Speedup} \\
\midrule
B200 & $4{\times}4096$ & 0.92\,{\tiny 1499} & \textbf{0.86}\,{\tiny 1600} & 1.09\,{\tiny 1263} & 1.12\,{\tiny 1228} & 1.25\,{\tiny 1103} & 1.34\,{\tiny 1028} & 4.27\,{\tiny 322} & 1.27$\times$ \\
 & $16{\times}4096$ & 3.87\,{\tiny 1420} & \textbf{3.69}\,{\tiny 1488} & 4.44\,{\tiny 1238} & 4.71\,{\tiny 1167} & 5.07\,{\tiny 1085} & 5.50\,{\tiny 999} & 16.8\,{\tiny 328} & 1.20$\times$ \\
 & $16{\times}1024$ & 0.31\,{\tiny 1114} & \textbf{0.30}\,{\tiny 1127} & 0.37\,{\tiny 937} & 0.47\,{\tiny 732} & 0.46\,{\tiny 756} & 0.40\,{\tiny 853} & 2.44\,{\tiny 141} & 1.20$\times$ \\
 & $64{\times}1024$ & 1.21\,{\tiny 1138} & \textbf{1.18}\,{\tiny 1167} & 1.38\,{\tiny 1000} & 1.86\,{\tiny 737} & 1.91\,{\tiny 721} & 1.74\,{\tiny 791} & 4.72\,{\tiny 291} & 1.17$\times$ \\
 & $8192 + 32{\times}256$ & 0.93\,{\tiny 1524} & \textbf{0.91}\,{\tiny 1564} & 1.21\,{\tiny 1175} & 1.19\,{\tiny 1188} & 1.30\,{\tiny 1089} & 1.53\,{\tiny 927} & 4.62\,{\tiny 307} & 1.32$\times$ \\
 & $8192 + 224{\times}256$ & 1.37\,{\tiny 1221} & \textbf{1.34}\,{\tiny 1246} & 2.48\,{\tiny 675} & 2.00\,{\tiny 836} & 2.44\,{\tiny 688} & 3.60\,{\tiny 465} & 6.86\,{\tiny 244} & 1.49$\times$ \\
 & log-unif.\ 16k & 0.67\,{\tiny 1347} & \textbf{0.65}\,{\tiny 1394} & 0.78\,{\tiny 1162} & 0.94\,{\tiny 960} & 0.87\,{\tiny 1037} & 0.93\,{\tiny 964} & 3.08\,{\tiny 292} & 1.20$\times$ \\
 & log-unif.\ 16k, GQA & 0.72\,{\tiny 1241} & \textbf{0.71}\,{\tiny 1264} & 0.91\,{\tiny 993} & {\tiny\texttt{no GQA}} & 0.93\,{\tiny 973} & 1.00\,{\tiny 901} & {\tiny\texttt{no GQA}} & 1.27$\times$ \\
 & log-unif.\ 64k & 3.95\,{\tiny 1385} & \textbf{3.78}\,{\tiny 1447} & 4.60\,{\tiny 1189} & 5.15\,{\tiny 1062} & 5.20\,{\tiny 1052} & 5.70\,{\tiny 959} & 17.2\,{\tiny 318} & 1.22$\times$ \\
\midrule
B300 & $4{\times}4096$ & 0.88\,{\tiny 1570} & \textbf{0.84}\,{\tiny 1626} & 0.85\,{\tiny 1613} & 1.05\,{\tiny 1304} & 1.13\,{\tiny 1139} & 1.32\,{\tiny 1088} & 4.26\,{\tiny 322} & 1.01$\times$ \\
 & $16{\times}4096$ & 3.55\,{\tiny 1533} & \textbf{3.43}\,{\tiny 1588} & 3.63\,{\tiny 1516} & 4.15\,{\tiny 1326} & 4.80\,{\tiny 1136} & 5.25\,{\tiny 1088} & 16.2\,{\tiny 340} & 1.06$\times$ \\
 & $16{\times}1024$ & 0.30\,{\tiny 1159} & \textbf{0.29}\,{\tiny 1183} & 0.32\,{\tiny 1079} & 0.45\,{\tiny 771} & 0.44\,{\tiny 776} & 0.38\,{\tiny 901} & 3.24\,{\tiny 88} & 1.09$\times$ \\
 & $64{\times}1024$ & 1.14\,{\tiny 1207} & \textbf{1.12}\,{\tiny 1224} & 1.19\,{\tiny 1152} & 1.75\,{\tiny 785} & 1.78\,{\tiny 798} & 1.68\,{\tiny 852} & 4.57\,{\tiny 295} & 1.06$\times$ \\
 & $8192 + 32{\times}256$ & 0.91\,{\tiny 1555} & \textbf{0.88}\,{\tiny 1613} & 0.99\,{\tiny 1438} & 1.08\,{\tiny 1308} & 1.17\,{\tiny 1235} & 1.43\,{\tiny 996} & 4.47\,{\tiny 316} & 1.12$\times$ \\
 & $8192 + 224{\times}256$ & 1.33\,{\tiny 1269} & \textbf{1.30}\,{\tiny 1295} & 2.25\,{\tiny 746} & 1.85\,{\tiny 905} & 2.27\,{\tiny 734} & 3.45\,{\tiny 486} & 6.50\,{\tiny 258} & 1.43$\times$ \\
 & log-unif.\ 16k & 0.65\,{\tiny 1376} & \textbf{0.63}\,{\tiny 1425} & 0.65\,{\tiny 1376} & 0.89\,{\tiny 1015} & 0.88\,{\tiny 1021} & 0.88\,{\tiny 1012} & 3.83\,{\tiny 212} & 1.03$\times$ \\
 & log-unif.\ 16k, GQA & 0.68\,{\tiny 1335} & \textbf{0.67}\,{\tiny 1338} & 0.79\,{\tiny 1133} & {\tiny\texttt{no GQA}} & 0.84\,{\tiny 1073} & 0.90\,{\tiny 1043} & {\tiny\texttt{no GQA}} & 1.17$\times$ \\
 & log-unif.\ 64k & 3.66\,{\tiny 1506} & \textbf{3.53}\,{\tiny 1548} & 3.77\,{\tiny 1448} & 4.70\,{\tiny 1166} & 4.99\,{\tiny 1097} & 5.42\,{\tiny 1042} & 16.6\,{\tiny 330} & 1.07$\times$ \\
\bottomrule
\end{tabular}
\end{adjustbox}
\end{table}

\section{Kernels on B300}
\label{app:b300}

\Cref{tab:varlen_fwd_b300,tab:varlen_bwd_b300} repeat the kernel measurements of \Cref{sec:eval_varlen} on B300.

\begin{table}[!htbp]
\centering
\caption{B300. Forward kernel on packed documents, causal, 16 heads, $d = 128$: time in ms (fastest in bold), TFLOP/s in small type; Speedup: the fastest baseline's time over ours. cuDNN MX: cuDNN MXFP8; FlashInfer: its fastest backend;}\label{tab:varlen_fwd_b300}
\scriptsize
\setlength{\tabcolsep}{1pt}
\renewcommand{\arraystretch}{1.0}
\begin{adjustbox}{max width=\linewidth,max totalheight=0.78\textheight}
\begin{tabular}{@{}llllllllllllll@{}}
\toprule
& & \multicolumn{2}{l}{\textbf{Ours}} & \multicolumn{4}{l}{\textbf{8-bit}} & \textbf{4-bit} & \multicolumn{4}{l}{\textbf{BF16}} & \\
\cmidrule(lr){3-4}\cmidrule(lr){5-8}\cmidrule(lr){9-9}\cmidrule(lr){10-13}
\textbf{GPU} & \textbf{Documents} & MXFP8 & NVFP4 & \shortstack{cuDNN\\FP8} & \shortstack{cuDNN\\MX} & FA4 FP8 & FlashInfer & FlashInfer & FA4 & cuDNN & FlashInfer & SDPA & \textbf{Speedup} \\
\midrule
B300 & $4{\times}4096$ & \shortstack{\textbf{0.14}\\{\tiny 1950}} & \shortstack{0.18\\{\tiny 1563}} & \shortstack{0.18\\{\tiny 1512}} & \shortstack{0.15\\{\tiny 1778}} & \shortstack{0.22\\{\tiny 1232}} & \shortstack{0.20\\{\tiny 1382}} & \shortstack{0.21\\{\tiny 1284}} & \shortstack{0.22\\{\tiny 1262}} & \shortstack{0.38\\{\tiny 723}} & \shortstack{0.24\\{\tiny 1140}} & \shortstack{2.98\\{\tiny 92}} & 1.10$\times$ \\
 & $16{\times}4096$ & \shortstack{0.56\\{\tiny 1956}} & \shortstack{0.67\\{\tiny 1640}} & \shortstack{\textbf{0.55}\\{\tiny 1992}} & \shortstack{0.57\\{\tiny 1921}} & \shortstack{0.87\\{\tiny 1270}} & \shortstack{0.71\\{\tiny 1541}} & \shortstack{0.75\\{\tiny 1466}} & \shortstack{0.91\\{\tiny 1154}} & \shortstack{1.38\\{\tiny 795}} & \shortstack{0.96\\{\tiny 1170}} & \shortstack{4.34\\{\tiny 257}} & 0.98$\times$ \\
 & $16{\times}1024$ & \shortstack{\textbf{0.06}\\{\tiny 1170}} & \shortstack{0.07\\{\tiny 1042}} & \shortstack{0.09\\{\tiny 761}} & \shortstack{0.08\\{\tiny 900}} & \shortstack{0.10\\{\tiny 689}} & \shortstack{0.13\\{\tiny 518}} & \shortstack{0.09\\{\tiny 798}} & \shortstack{0.11\\{\tiny 651}} & \shortstack{0.21\\{\tiny 322}} & \shortstack{0.14\\{\tiny 487}} & \shortstack{2.97\\{\tiny 23}} & 1.30$\times$ \\
 & $64{\times}1024$ & \shortstack{\textbf{0.21}\\{\tiny 1303}} & \shortstack{0.23\\{\tiny 1195}} & \shortstack{0.28\\{\tiny 999}} & \shortstack{0.26\\{\tiny 1070}} & \shortstack{0.39\\{\tiny 708}} & \shortstack{0.52\\{\tiny 529}} & \shortstack{0.30\\{\tiny 911}} & \shortstack{0.41\\{\tiny 665}} & \shortstack{1.01\\{\tiny 271}} & \shortstack{0.55\\{\tiny 503}} & \shortstack{2.74\\{\tiny 91}} & 1.22$\times$ \\
 & $8192 + 32{\times}256$ & \shortstack{\textbf{0.14}\\{\tiny 1956}} & \shortstack{0.18\\{\tiny 1563}} & \shortstack{0.22\\{\tiny 1299}} & \shortstack{0.43\\{\tiny 660}} & \shortstack{0.23\\{\tiny 1238}} & \shortstack{0.33\\{\tiny 858}} & \shortstack{0.26\\{\tiny 1106}} & \shortstack{0.22\\{\tiny 1294}} & \shortstack{0.40\\{\tiny 716}} & \shortstack{0.37\\{\tiny 769}} & \shortstack{2.56\\{\tiny 111}} & 1.51$\times$ \\
 & $8192 + 224{\times}256$ & \shortstack{\textbf{0.24}\\{\tiny 1419}} & \shortstack{0.28\\{\tiny 1203}} & \shortstack{0.40\\{\tiny 840}} & \shortstack{2.19\\{\tiny 153}} & \shortstack{0.52\\{\tiny 646}} & \shortstack{0.60\\{\tiny 558}} & \shortstack{0.75\\{\tiny 444}} & \shortstack{0.53\\{\tiny 635}} & \shortstack{3.47\\{\tiny 97}} & \shortstack{0.69\\{\tiny 483}} & \shortstack{2.99\\{\tiny 110}} & 1.69$\times$ \\
 & log-unif.\ 16k & \shortstack{\textbf{0.11}\\{\tiny 1634}} & \shortstack{0.14\\{\tiny 1331}} & \shortstack{0.15\\{\tiny 1176}} & \shortstack{0.16\\{\tiny 1102}} & \shortstack{0.18\\{\tiny 1002}} & \shortstack{0.18\\{\tiny 973}} & \shortstack{0.17\\{\tiny 1082}} & \shortstack{0.18\\{\tiny 1019}} & \shortstack{0.33\\{\tiny 564}} & \shortstack{0.21\\{\tiny 847}} & \shortstack{3.02\\{\tiny 57}} & 1.39$\times$ \\
 & log-unif.\ 16k, GQA & \shortstack{\textbf{0.11}\\{\tiny 1649}} & \shortstack{0.13\\{\tiny 1344}} & \shortstack{0.15\\{\tiny 1161}} & \shortstack{0.16\\{\tiny 1116}} & \shortstack{0.15\\{\tiny 1174}} & \shortstack{0.30\\{\tiny 595}} & \shortstack{0.16\\{\tiny 1092}} & \shortstack{0.15\\{\tiny 1225}} & \shortstack{0.31\\{\tiny 587}} & \shortstack{0.35\\{\tiny 517}} & {\tiny\texttt{no GQA}} & 1.35$\times$ \\
 & log-unif.\ 64k & \shortstack{\textbf{0.58}\\{\tiny 1919}} & \shortstack{0.70\\{\tiny 1556}} & \shortstack{0.60\\{\tiny 1827}} & \shortstack{0.78\\{\tiny 1401}} & \shortstack{0.90\\{\tiny 1218}} & \shortstack{0.81\\{\tiny 1359}} & \shortstack{0.81\\{\tiny 1351}} & \shortstack{0.99\\{\tiny 1262}} & \shortstack{1.56\\{\tiny 702}} & \shortstack{1.06\\{\tiny 1032}} & \shortstack{4.55\\{\tiny 240}} & 1.05$\times$ \\
\bottomrule
\end{tabular}
\end{adjustbox}
\end{table}

\begin{table}[!htbp]
\centering
\caption{B300. Backward kernels on packed documents, causal, 16 heads, $d = 128$: time in ms (fastest in bold), TFLOP/s in small type; Speedup: the fastest baseline's time over ours.}\label{tab:varlen_bwd_b300}
\scriptsize
\setlength{\tabcolsep}{1.5pt}
\renewcommand{\arraystretch}{1.0}
\begin{adjustbox}{max width=\linewidth,max totalheight=0.78\textheight}
\begin{tabular}{@{}lllllllll@{}}
\toprule
& & \multicolumn{2}{l}{\textbf{Ours}} & \textbf{FP8} & \multicolumn{3}{l}{\textbf{BF16}} & \\
\cmidrule(lr){3-4}\cmidrule(lr){6-8}
\textbf{GPU} & \textbf{Documents} & MXFP8 & NVFP4 & cuDNN & FA4 & cuDNN & SDPA & \textbf{Speedup} \\
\midrule
B300 & $4{\times}4096$ & 0.51\,{\tiny 1351} & \textbf{0.49}\,{\tiny 1404} & 0.61\,{\tiny 1121} & 0.66\,{\tiny 1048} & 0.71\,{\tiny 963} & 3.96\,{\tiny 164} & 1.25$\times$ \\
 & $16{\times}4096$ & 2.04\,{\tiny 1347} & \textbf{1.96}\,{\tiny 1413} & 2.30\,{\tiny 1196} & 2.90\,{\tiny 948} & 3.10\,{\tiny 896} & 8.90\,{\tiny 308} & 1.18$\times$ \\
 & $16{\times}1024$ & 0.22\,{\tiny 783} & \textbf{0.22}\,{\tiny 799} & 0.27\,{\tiny 630} & 0.35\,{\tiny 494} & 0.30\,{\tiny 573} & 3.85\,{\tiny 44} & 1.26$\times$ \\
 & $64{\times}1024$ & 0.84\,{\tiny 820} & \textbf{0.83}\,{\tiny 829} & 0.96\,{\tiny 713} & 1.38\,{\tiny 499} & 1.12\,{\tiny 620} & 3.68\,{\tiny 175} & 1.17$\times$ \\
 & $8192 + 32{\times}256$ & 0.52\,{\tiny 1363} & \textbf{0.50}\,{\tiny 1419} & 0.74\,{\tiny 954} & 0.67\,{\tiny 1055} & 0.99\,{\tiny 715} & 3.71\,{\tiny 164} & 1.34$\times$ \\
 & $8192 + 224{\times}256$ & 0.93\,{\tiny 900} & \textbf{0.92}\,{\tiny 913} & 1.99\,{\tiny 422} & 1.83\,{\tiny 457} & 3.00\,{\tiny 279} & 4.69\,{\tiny 180} & 2.00$\times$ \\
 & log-unif.\ 16k & 0.40\,{\tiny 1128} & \textbf{0.39}\,{\tiny 1166} & 0.51\,{\tiny 881} & 0.53\,{\tiny 847} & 0.55\,{\tiny 816} & 3.39\,{\tiny 133} & 1.33$\times$ \\
 & log-unif.\ 16k, GQA & 0.45\,{\tiny 992} & \textbf{0.44}\,{\tiny 1016} & 0.64\,{\tiny 705} & 0.56\,{\tiny 802} & 0.58\,{\tiny 770} & {\tiny\texttt{no GQA}} & 1.27$\times$ \\
 & log-unif.\ 64k & 2.09\,{\tiny 1312} & \textbf{2.01}\,{\tiny 1366} & 2.49\,{\tiny 1102} & 2.98\,{\tiny 924} & 3.21\,{\tiny 863} & 9.16\,{\tiny 298} & 1.24$\times$ \\
\bottomrule
\end{tabular}
\end{adjustbox}
\end{table}

\section{Batched Attention}
\label{app:batched}

\Cref{tab:batched_train} gives the training call on batched sequences with the baselines of \Cref{sec:experiments}. cuDNN FP8 needs its scales before the call. Over 30 packed and batched shapes, our training call is faster than cuDNN FP8 by a median of 7.4\,\% when every cuDNN scale is calibrated beforehand, 10.3\,\% when its input scales are computed in the call, and 112\,\% when all of its scales are, because computing every scale requires a second forward and backward pass. In the inference call on batched sequences, Schraudolph's exponential makes our MXFP8 route 7--26\,\% faster and ahead of cuDNN FP8 on every shape, and at $1{\times}32768$ non-causal, Mitchell-P takes 4.64\,ms against 6.17\,ms for our NVFP4 route and 4.68\,ms for the fastest 8-bit baseline.

\begin{table}[!htbp]
\centering
\caption{Training call on batched sequences (forward and backward, quantization included): time in ms (fastest in bold), TFLOP/s in small type. cuDNN FP8 with every scale calibrated beforehand, with its input scales computed in the call, and with every scale computed in the call. Speedup: the fastest baseline's time, with cuDNN FP8 computing its input scales, over ours.}\label{tab:batched_train}
\scriptsize
\setlength{\tabcolsep}{2pt}
\renewcommand{\arraystretch}{1.0}
\begin{adjustbox}{max width=\linewidth,max totalheight=0.78\textheight}
\begin{tabular}{@{}llllllllllll@{}}
\toprule
& & \multicolumn{2}{l}{\textbf{Ours}} & \multicolumn{4}{l}{\textbf{cuDNN}} & \textbf{LP-FA4} & \textbf{FA4} & \textbf{cuDNN} & \\
\cmidrule(lr){3-4}\cmidrule(lr){5-8}
\textbf{GPU} & \textbf{Shape} & MXFP8 & NVFP4 & FP8 cal. & FP8 fresh in. & FP8 all fresh & MXFP8 & MXFP8 & BF16 & BF16 & \textbf{Speedup} \\
\midrule
B200 & $4{\times}4096$, 16 h, $d{=}128$, causal & \shortstack{\textbf{0.87}\\{\tiny 1108}} & \shortstack{0.87\\{\tiny 1105}} & \shortstack{0.88\\{\tiny 1096}} & \shortstack{0.91\\{\tiny 1059}} & \shortstack{1.74\\{\tiny 553}} & \shortstack{1.25\\{\tiny 767}} & {\tiny\texttt{no causal}} & \shortstack{1.05\\{\tiny 918}} & \shortstack{1.05\\{\tiny 916}} & 1.05$\times$ \\
 & $4{\times}4096$, 16 h, $d{=}128$, non-causal & \shortstack{1.41\\{\tiny 1363}} & \shortstack{\textbf{1.41}\\{\tiny 1364}} & \shortstack{\textbf{1.41}\\{\tiny 1364}} & \shortstack{1.44\\{\tiny 1337}} & \shortstack{2.79\\{\tiny 689}} & \shortstack{1.85\\{\tiny 1037}} & \shortstack{6.12\\{\tiny 315}} & \shortstack{1.68\\{\tiny 1148}} & \shortstack{1.74\\{\tiny 1106}} & 1.02$\times$ \\
 & $1{\times}32768$, 16 h, $d{=}128$, causal & \shortstack{\textbf{9.88}\\{\tiny 1558}} & \shortstack{9.99\\{\tiny 1541}} & \shortstack{10.6\\{\tiny 1447}} & \shortstack{10.7\\{\tiny 1440}} & \shortstack{21.1\\{\tiny 730}} & \shortstack{12.6\\{\tiny 1223}} & {\tiny\texttt{no causal}} & \shortstack{12.6\\{\tiny 1224}} & \shortstack{13.4\\{\tiny 1145}} & 1.08$\times$ \\
 & $1{\times}32768$, 16 h, $d{=}128$, non-causal & \shortstack{19.1\\{\tiny 1615}} & \shortstack{\textbf{19.0}\\{\tiny 1617}} & \shortstack{20.0\\{\tiny 1540}} & \shortstack{20.0\\{\tiny 1540}} & \shortstack{39.6\\{\tiny 778}} & \shortstack{23.9\\{\tiny 1288}} & \shortstack{28.6\\{\tiny 1077}} & \shortstack{23.8\\{\tiny 1294}} & \shortstack{25.5\\{\tiny 1208}} & 1.05$\times$ \\
 & $2{\times}8192$, 16 h, $d{=}128$, causal & \shortstack{1.45\\{\tiny 1326}} & \shortstack{\textbf{1.45}\\{\tiny 1327}} & \shortstack{1.53\\{\tiny 1259}} & \shortstack{1.56\\{\tiny 1231}} & \shortstack{3.03\\{\tiny 636}} & \shortstack{2.03\\{\tiny 948}} & {\tiny\texttt{no causal}} & \shortstack{1.79\\{\tiny 1073}} & \shortstack{1.85\\{\tiny 1040}} & 1.08$\times$ \\
 & $2{\times}8192$, 16 h, $d{=}128$, non-causal & \shortstack{2.59\\{\tiny 1484}} & \shortstack{\textbf{2.56}\\{\tiny 1503}} & \shortstack{2.63\\{\tiny 1463}} & \shortstack{2.67\\{\tiny 1444}} & \shortstack{5.22\\{\tiny 738}} & \shortstack{3.32\\{\tiny 1160}} & \shortstack{6.99\\{\tiny 551}} & \shortstack{3.13\\{\tiny 1231}} & \shortstack{3.31\\{\tiny 1164}} & 1.04$\times$ \\
 & $1{\times}16384$, 16 h, $d{=}128$, causal & \shortstack{2.64\\{\tiny 1456}} & \shortstack{\textbf{2.61}\\{\tiny 1473}} & \shortstack{2.81\\{\tiny 1370}} & \shortstack{2.85\\{\tiny 1351}} & \shortstack{5.58\\{\tiny 690}} & \shortstack{3.54\\{\tiny 1088}} & {\tiny\texttt{no causal}} & \shortstack{3.34\\{\tiny 1154}} & \shortstack{3.55\\{\tiny 1083}} & 1.09$\times$ \\
 & $1{\times}16384$, 16 h, $d{=}128$, non-causal & \shortstack{4.94\\{\tiny 1558}} & \shortstack{\textbf{4.87}\\{\tiny 1581}} & \shortstack{5.06\\{\tiny 1520}} & \shortstack{5.09\\{\tiny 1513}} & \shortstack{10.0\\{\tiny 766}} & \shortstack{6.20\\{\tiny 1242}} & \shortstack{9.29\\{\tiny 828}} & \shortstack{6.16\\{\tiny 1249}} & \shortstack{6.57\\{\tiny 1172}} & 1.05$\times$ \\
 & $4{\times}4096$, 16/4 h, $d{=}128$, causal & \shortstack{\textbf{0.80}\\{\tiny 1200}} & \shortstack{0.82\\{\tiny 1178}} & \shortstack{0.96\\{\tiny 1005}} & \shortstack{0.98\\{\tiny 985}} & \shortstack{1.90\\{\tiny 506}} & \shortstack{1.24\\{\tiny 776}} & {\tiny\texttt{no causal}} & \shortstack{1.08\\{\tiny 892}} & \shortstack{1.04\\{\tiny 924}} & 1.22$\times$ \\
 & $4{\times}4096$, 16/4 h, $d{=}128$, non-causal & \shortstack{1.44\\{\tiny 1336}} & \shortstack{\textbf{1.43}\\{\tiny 1342}} & \shortstack{1.50\\{\tiny 1285}} & \shortstack{1.52\\{\tiny 1268}} & \shortstack{2.98\\{\tiny 646}} & \shortstack{1.86\\{\tiny 1035}} & {\tiny\texttt{no GQA}} & \shortstack{1.76\\{\tiny 1093}} & \shortstack{1.78\\{\tiny 1083}} & 1.06$\times$ \\
 & $4{\times}4096$, 16 h, $d{=}64$, causal & \shortstack{\textbf{0.65}\\{\tiny 740}} & \shortstack{0.70\\{\tiny 692}} & \shortstack{0.72\\{\tiny 664}} & \shortstack{0.74\\{\tiny 647}} & \shortstack{1.47\\{\tiny 329}} & {\tiny\texttt{no d=64}} & {\tiny\texttt{no causal}} & \shortstack{0.72\\{\tiny 670}} & \shortstack{0.76\\{\tiny 633}} & 1.10$\times$ \\
 & $4{\times}4096$, 16 h, $d{=}64$, non-causal & \shortstack{\textbf{1.08}\\{\tiny 887}} & \shortstack{1.15\\{\tiny 836}} & \shortstack{1.19\\{\tiny 806}} & \shortstack{1.21\\{\tiny 793}} & \shortstack{2.40\\{\tiny 400}} & {\tiny\texttt{no d=64}} & {\tiny\texttt{no d=64}} & \shortstack{1.20\\{\tiny 799}} & \shortstack{1.28\\{\tiny 752}} & 1.11$\times$ \\
\midrule
B300 & $4{\times}4096$, 16 h, $d{=}128$, causal & \shortstack{0.79\\{\tiny 1220}} & \shortstack{0.80\\{\tiny 1208}} & \shortstack{\textbf{0.73}\\{\tiny 1316}} & \shortstack{0.75\\{\tiny 1279}} & \shortstack{1.42\\{\tiny 676}} & \shortstack{1.13\\{\tiny 853}} & {\tiny\texttt{no causal}} & \shortstack{0.95\\{\tiny 1012}} & \shortstack{0.97\\{\tiny 996}} & 0.95$\times$ \\
 & $4{\times}4096$, 16 h, $d{=}128$, non-causal & \shortstack{1.20\\{\tiny 1558}} & \shortstack{1.25\\{\tiny 1540}} & \shortstack{\textbf{1.04}\\{\tiny 1686}} & \shortstack{1.07\\{\tiny 1805}} & \shortstack{2.05\\{\tiny 860}} & \shortstack{1.49\\{\tiny 1293}} & \shortstack{5.75\\{\tiny 335}} & \shortstack{1.38\\{\tiny 1454}} & \shortstack{1.38\\{\tiny 1394}} & 0.89$\times$ \\
 & $1{\times}32768$, 16 h, $d{=}128$, causal & \shortstack{8.68\\{\tiny 1774}} & \shortstack{8.89\\{\tiny 1731}} & \shortstack{\textbf{8.49}\\{\tiny 1813}} & \shortstack{8.57\\{\tiny 1810}} & \shortstack{16.7\\{\tiny 920}} & \shortstack{10.8\\{\tiny 1420}} & {\tiny\texttt{no causal}} & \shortstack{11.7\\{\tiny 1316}} & \shortstack{12.5\\{\tiny 1232}} & 0.99$\times$ \\
 & $1{\times}32768$, 16 h, $d{=}128$, non-causal & \shortstack{15.8\\{\tiny 1879}} & \shortstack{16.4\\{\tiny 1850}} & \shortstack{16.5\\{\tiny 1792}} & \shortstack{\textbf{15.0}\\{\tiny 2096}} & \shortstack{34.0\\{\tiny 935}} & \shortstack{19.3\\{\tiny 1742}} & \shortstack{25.4\\{\tiny 1212}} & \shortstack{22.6\\{\tiny 1361}} & \shortstack{23.0\\{\tiny 1180}} & 0.95$\times$ \\
 & $2{\times}8192$, 16 h, $d{=}128$, causal & \shortstack{1.27\\{\tiny 1520}} & \shortstack{1.31\\{\tiny 1467}} & \shortstack{\textbf{1.11}\\{\tiny 1733}} & \shortstack{1.14\\{\tiny 1690}} & \shortstack{2.20\\{\tiny 875}} & \shortstack{1.73\\{\tiny 1111}} & {\tiny\texttt{no causal}} & \shortstack{1.41\\{\tiny 1363}} & \shortstack{1.48\\{\tiny 1303}} & 0.90$\times$ \\
 & $2{\times}8192$, 16 h, $d{=}128$, non-causal & \shortstack{2.17\\{\tiny 1777}} & \shortstack{2.24\\{\tiny 1720}} & \shortstack{\textbf{1.86}\\{\tiny 2066}} & \shortstack{1.89\\{\tiny 1788}} & \shortstack{3.68\\{\tiny 1044}} & \shortstack{2.54\\{\tiny 1413}} & \shortstack{6.59\\{\tiny 592}} & \shortstack{2.43\\{\tiny 1586}} & \shortstack{2.61\\{\tiny 1473}} & 0.87$\times$ \\
 & $1{\times}16384$, 16 h, $d{=}128$, causal & \shortstack{2.37\\{\tiny 1626}} & \shortstack{2.37\\{\tiny 1623}} & \shortstack{\textbf{2.26}\\{\tiny 1702}} & \shortstack{2.28\\{\tiny 1691}} & \shortstack{4.44\\{\tiny 867}} & \shortstack{3.08\\{\tiny 1252}} & {\tiny\texttt{no causal}} & \shortstack{3.07\\{\tiny 1255}} & \shortstack{3.30\\{\tiny 1167}} & 0.96$\times$ \\
 & $1{\times}16384$, 16 h, $d{=}128$, non-causal & \shortstack{4.33\\{\tiny 1775}} & \shortstack{4.31\\{\tiny 1787}} & \shortstack{\textbf{4.16}\\{\tiny 1851}} & \shortstack{4.18\\{\tiny 1840}} & \shortstack{8.17\\{\tiny 941}} & \shortstack{5.13\\{\tiny 1500}} & \shortstack{8.58\\{\tiny 896}} & \shortstack{5.71\\{\tiny 1348}} & \shortstack{6.09\\{\tiny 1264}} & 0.97$\times$ \\
 & $4{\times}4096$, 16/4 h, $d{=}128$, causal & \shortstack{\textbf{0.72}\\{\tiny 1338}} & \shortstack{0.74\\{\tiny 1308}} & \shortstack{0.80\\{\tiny 1210}} & \shortstack{0.81\\{\tiny 1184}} & \shortstack{1.58\\{\tiny 608}} & \shortstack{1.12\\{\tiny 857}} & {\tiny\texttt{no causal}} & \shortstack{0.97\\{\tiny 990}} & \shortstack{0.96\\{\tiny 1001}} & 1.13$\times$ \\
 & $4{\times}4096$, 16/4 h, $d{=}128$, non-causal & \shortstack{1.26\\{\tiny 1530}} & \shortstack{1.29\\{\tiny 1486}} & \shortstack{\textbf{1.25}\\{\tiny 1546}} & \shortstack{1.26\\{\tiny 1525}} & \shortstack{2.46\\{\tiny 781}} & \shortstack{1.57\\{\tiny 1222}} & {\tiny\texttt{no GQA}} & \shortstack{1.61\\{\tiny 1192}} & \shortstack{1.65\\{\tiny 1165}} & 1.00$\times$ \\
 & $4{\times}4096$, 16 h, $d{=}64$, causal & \shortstack{\textbf{0.58}\\{\tiny 830}} & \shortstack{0.65\\{\tiny 744}} & \shortstack{0.58\\{\tiny 827}} & \shortstack{0.59\\{\tiny 809}} & \shortstack{1.16\\{\tiny 416}} & {\tiny\texttt{no d=64}} & {\tiny\texttt{no causal}} & \shortstack{0.61\\{\tiny 774}} & \shortstack{0.67\\{\tiny 710}} & 1.02$\times$ \\
 & $4{\times}4096$, 16 h, $d{=}64$, non-causal & \shortstack{0.94\\{\tiny 1020}} & \shortstack{1.00\\{\tiny 964}} & \shortstack{\textbf{0.92}\\{\tiny 1047}} & \shortstack{0.93\\{\tiny 1032}} & \shortstack{1.83\\{\tiny 526}} & {\tiny\texttt{no d=64}} & {\tiny\texttt{no d=64}} & \shortstack{1.01\\{\tiny 951}} & \shortstack{1.11\\{\tiny 831}} & 0.99$\times$ \\
\bottomrule
\end{tabular}
\end{adjustbox}
\end{table}

\Cref{tab:batched_fwd,tab:batched_bwd} give the forward and backward kernels alone on the same shapes, with operands quantized outside the timed region. Our backward is the fastest on all twelve B200 rows and on six of twelve B300 rows, where cuDNN FP8 leads the other six by 3--8\,\%. Our forward leads on three B200 rows and on no B300 row. cuDNN FP8 is faster on most rows, and without a causal mask the 4-bit Direct-P forward of HW-FP4 FA4 is the fastest on every MHA row with $d = 128$.

\begin{table}[!htbp]
\centering
\caption{Forward kernel on batched sequences, kernels alone with operands quantized outside the timed region: time in ms (fastest in bold), TFLOP/s in small type. FlashInfer: its fastest backend, FlashInfer 4-bit: NVFP4 KV cache. Speedup: the fastest baseline's time over ours.}\label{tab:batched_fwd}
\scriptsize
\setlength{\tabcolsep}{1.5pt}
\renewcommand{\arraystretch}{1.0}
\begin{adjustbox}{max width=\linewidth,max totalheight=0.87\textheight}
\begin{tabular}{@{}lp{3.6cm}lllllllllllllll@{}}
\toprule
 & & \multicolumn{2}{l}{\textbf{Ours}} & \multicolumn{2}{l}{\textbf{4-bit}} & \multicolumn{5}{l}{\textbf{8-bit}} & \multicolumn{5}{l}{\textbf{BF16}} & \\
\cmidrule(lr){3-4}\cmidrule(lr){5-6}\cmidrule(lr){7-11}\cmidrule(lr){12-16}
\textbf{GPU} & \textbf{Shape} & {\tiny\shortstack[l]{MXFP8}} & {\tiny\shortstack[l]{NVFP4}} & {\tiny\shortstack[l]{Direct-P}} & {\tiny\shortstack[l]{FlashInfer}} & {\tiny\shortstack[l]{cuDNN\\FP8}} & {\tiny\shortstack[l]{cuDNN\\MX}} & {\tiny\shortstack[l]{FA4\\FP8}} & {\tiny\shortstack[l]{FlashInfer}} & {\tiny\shortstack[l]{LP-FA4}} & {\tiny\shortstack[l]{FA4}} & {\tiny\shortstack[l]{cuDNN}} & {\tiny\shortstack[l]{FlashInfer}} & {\tiny\shortstack[l]{SDPA}} & {\tiny\shortstack[l]{LP-FA4}} & \textbf{Speedup} \\
\midrule
B200 & \raggedright $4{\times}4096$, 16 h, $d{=}128$, causal & \shortstack{\textbf{0.17}\\{\tiny 1662}} & \shortstack{0.21\\{\tiny 1283}} & \shortstack{{\tiny\texttt{no}}\\{\tiny\texttt{causal}}} & \shortstack{0.24\\{\tiny 1142}} & \shortstack{0.18\\{\tiny 1551}} & \shortstack{0.32\\{\tiny 853}} & \shortstack{0.20\\{\tiny 1341}} & \shortstack{0.23\\{\tiny 1211}} & \shortstack{{\tiny\texttt{no}}\\{\tiny\texttt{causal}}} & \shortstack{0.23\\{\tiny 1210}} & \shortstack{0.28\\{\tiny 976}} & \shortstack{0.25\\{\tiny 1087}} & \shortstack{0.22\\{\tiny 1268}} & \shortstack{0.23\\{\tiny 1209}} & 1.07$\times$ \\
 & \raggedright $4{\times}4096$, 16 h, $d{=}128$, non-causal & \shortstack{0.30\\{\tiny 1813}} & \shortstack{0.40\\{\tiny 1392}} & \shortstack{\textbf{0.20}\\{\tiny 2677}} & \shortstack{0.36\\{\tiny 1517}} & \shortstack{0.29\\{\tiny 1878}} & \shortstack{0.39\\{\tiny 1426}} & \shortstack{0.35\\{\tiny 1547}} & \shortstack{0.29\\{\tiny 1902}} & \shortstack{0.32\\{\tiny 1716}} & \shortstack{0.37\\{\tiny 1490}} & \shortstack{0.39\\{\tiny 1397}} & \shortstack{0.40\\{\tiny 1391}} & \shortstack{0.37\\{\tiny 1502}} & \shortstack{0.50\\{\tiny 1102}} & 0.68$\times$ \\
 & \raggedright $1{\times}32768$, 16 h, $d{=}128$, causal & \shortstack{2.54\\{\tiny 1729}} & \shortstack{3.04\\{\tiny 1449}} & \shortstack{{\tiny\texttt{no}}\\{\tiny\texttt{causal}}} & \shortstack{2.88\\{\tiny 1530}} & \shortstack{\textbf{2.49}\\{\tiny 1765}} & \shortstack{3.06\\{\tiny 1435}} & \shortstack{2.57\\{\tiny 1714}} & \shortstack{2.55\\{\tiny 1728}} & \shortstack{{\tiny\texttt{no}}\\{\tiny\texttt{causal}}} & \shortstack{3.26\\{\tiny 1349}} & \shortstack{3.71\\{\tiny 1184}} & \shortstack{3.23\\{\tiny 1360}} & \shortstack{3.26\\{\tiny 1350}} & \shortstack{3.27\\{\tiny 1346}} & 0.98$\times$ \\
 & \raggedright $1{\times}32768$, 16 h, $d{=}128$, non-causal & \shortstack{5.15\\{\tiny 1708}} & \shortstack{6.00\\{\tiny 1466}} & \shortstack{\textbf{3.11}\\{\tiny 2832}} & \shortstack{5.36\\{\tiny 1642}} & \shortstack{4.85\\{\tiny 1814}} & \shortstack{5.61\\{\tiny 1569}} & \shortstack{5.28\\{\tiny 1668}} & \shortstack{4.60\\{\tiny 1914}} & \shortstack{5.20\\{\tiny 1692}} & \shortstack{6.34\\{\tiny 1387}} & \shortstack{6.81\\{\tiny 1292}} & \shortstack{5.91\\{\tiny 1490}} & \shortstack{6.21\\{\tiny 1415}} & \shortstack{8.47\\{\tiny 1038}} & 0.60$\times$ \\
 & \raggedright $2{\times}8192$, 16 h, $d{=}128$, causal & \shortstack{\textbf{0.31}\\{\tiny 1785}} & \shortstack{0.41\\{\tiny 1358}} & \shortstack{{\tiny\texttt{no}}\\{\tiny\texttt{causal}}} & \shortstack{0.43\\{\tiny 1274}} & \shortstack{0.31\\{\tiny 1778}} & \shortstack{0.54\\{\tiny 1018}} & \shortstack{0.35\\{\tiny 1580}} & \shortstack{0.37\\{\tiny 1481}} & \shortstack{{\tiny\texttt{no}}\\{\tiny\texttt{causal}}} & \shortstack{0.40\\{\tiny 1391}} & \shortstack{0.50\\{\tiny 1097}} & \shortstack{0.43\\{\tiny 1275}} & \shortstack{0.39\\{\tiny 1426}} & \shortstack{0.40\\{\tiny 1390}} & 1.00$\times$ \\
 & \raggedright $2{\times}8192$, 16 h, $d{=}128$, non-causal & \shortstack{0.60\\{\tiny 1833}} & \shortstack{0.77\\{\tiny 1425}} & \shortstack{\textbf{0.40}\\{\tiny 2767}} & \shortstack{0.69\\{\tiny 1605}} & \shortstack{0.56\\{\tiny 1955}} & \shortstack{0.73\\{\tiny 1502}} & \shortstack{0.69\\{\tiny 1595}} & \shortstack{0.54\\{\tiny 2039}} & \shortstack{0.67\\{\tiny 1631}} & \shortstack{0.73\\{\tiny 1505}} & \shortstack{0.80\\{\tiny 1381}} & \shortstack{0.78\\{\tiny 1410}} & \shortstack{0.72\\{\tiny 1525}} & \shortstack{1.01\\{\tiny 1088}} & 0.66$\times$ \\
 & \raggedright $1{\times}16384$, 16 h, $d{=}128$, causal & \shortstack{0.59\\{\tiny 1867}} & \shortstack{0.79\\{\tiny 1401}} & \shortstack{{\tiny\texttt{no}}\\{\tiny\texttt{causal}}} & \shortstack{0.81\\{\tiny 1365}} & \shortstack{\textbf{0.58}\\{\tiny 1896}} & \shortstack{0.98\\{\tiny 1125}} & \shortstack{0.64\\{\tiny 1732}} & \shortstack{0.65\\{\tiny 1688}} & \shortstack{{\tiny\texttt{no}}\\{\tiny\texttt{causal}}} & \shortstack{0.78\\{\tiny 1409}} & \shortstack{1.01\\{\tiny 1091}} & \shortstack{0.83\\{\tiny 1323}} & \shortstack{0.75\\{\tiny 1471}} & \shortstack{0.76\\{\tiny 1456}} & 0.98$\times$ \\
 & \raggedright $1{\times}16384$, 16 h, $d{=}128$, non-causal & \shortstack{1.25\\{\tiny 1761}} & \shortstack{1.52\\{\tiny 1448}} & \shortstack{\textbf{0.79}\\{\tiny 2802}} & \shortstack{1.34\\{\tiny 1636}} & \shortstack{1.22\\{\tiny 1806}} & \shortstack{1.43\\{\tiny 1543}} & \shortstack{1.34\\{\tiny 1636}} & \shortstack{1.08\\{\tiny 2030}} & \shortstack{1.32\\{\tiny 1663}} & \shortstack{1.59\\{\tiny 1387}} & \shortstack{1.71\\{\tiny 1284}} & \shortstack{1.59\\{\tiny 1378}} & \shortstack{1.56\\{\tiny 1406}} & \shortstack{2.14\\{\tiny 1029}} & 0.63$\times$ \\
 & \raggedright $4{\times}4096$, 16/4 h, $d{=}128$, causal & \shortstack{\textbf{0.16}\\{\tiny 1676}} & \shortstack{0.21\\{\tiny 1284}} & \shortstack{{\tiny\texttt{no}}\\{\tiny\texttt{causal}}} & \shortstack{0.24\\{\tiny 1145}} & \shortstack{0.18\\{\tiny 1556}} & \shortstack{0.28\\{\tiny 970}} & \shortstack{0.19\\{\tiny 1431}} & \shortstack{0.23\\{\tiny 1222}} & \shortstack{{\tiny\texttt{no}}\\{\tiny\texttt{causal}}} & \shortstack{0.21\\{\tiny 1301}} & \shortstack{0.25\\{\tiny 1112}} & \shortstack{0.25\\{\tiny 1114}} & \shortstack{0.21\\{\tiny 1305}} & \shortstack{0.21\\{\tiny 1309}} & 1.08$\times$ \\
 & \raggedright $4{\times}4096$, 16/4 h, $d{=}128$, non-causal & \shortstack{0.30\\{\tiny 1814}} & \shortstack{0.40\\{\tiny 1393}} & \shortstack{{\tiny\texttt{no}}\\{\tiny\texttt{GQA}}} & \shortstack{0.36\\{\tiny 1526}} & \shortstack{0.29\\{\tiny 1881}} & \shortstack{0.39\\{\tiny 1427}} & \shortstack{0.35\\{\tiny 1558}} & \shortstack{\textbf{0.29}\\{\tiny 1900}} & \shortstack{{\tiny\texttt{no}}\\{\tiny\texttt{GQA}}} & \shortstack{0.36\\{\tiny 1525}} & \shortstack{0.40\\{\tiny 1391}} & \shortstack{0.38\\{\tiny 1464}} & \shortstack{0.37\\{\tiny 1498}} & \shortstack{0.49\\{\tiny 1123}} & 0.95$\times$ \\
 & \raggedright $4{\times}4096$, 16 h, $d{=}64$, causal & \shortstack{0.19\\{\tiny 737}} & \shortstack{0.22\\{\tiny 638}} & \shortstack{{\tiny\texttt{no}}\\{\tiny\texttt{causal}}} & \shortstack{0.21\\{\tiny 652}} & \shortstack{\textbf{0.17}\\{\tiny 809}} & \shortstack{{\tiny\texttt{no}}\\{\tiny\texttt{d=64}}} & \shortstack{0.18\\{\tiny 743}} & \shortstack{0.26\\{\tiny 532}} & \shortstack{{\tiny\texttt{no}}\\{\tiny\texttt{causal}}} & \shortstack{0.19\\{\tiny 719}} & \shortstack{0.28\\{\tiny 496}} & \shortstack{0.29\\{\tiny 473}} & \shortstack{\textbf{0.17}\\{\tiny 807}} & \shortstack{0.19\\{\tiny 710}} & 0.91$\times$ \\
 & \raggedright $4{\times}4096$, 16 h, $d{=}64$, non-causal & \shortstack{0.31\\{\tiny 882}} & \shortstack{0.37\\{\tiny 736}} & \shortstack{{\tiny\texttt{no}}\\{\tiny\texttt{d=64}}} & \shortstack{0.32\\{\tiny 848}} & \shortstack{0.29\\{\tiny 959}} & \shortstack{{\tiny\texttt{no}}\\{\tiny\texttt{d=64}}} & \shortstack{0.31\\{\tiny 895}} & \shortstack{0.36\\{\tiny 753}} & \shortstack{{\tiny\texttt{no}}\\{\tiny\texttt{d=64}}} & \shortstack{0.29\\{\tiny 944}} & \shortstack{0.37\\{\tiny 735}} & \shortstack{0.41\\{\tiny 673}} & \shortstack{\textbf{0.29}\\{\tiny 960}} & \shortstack{0.30\\{\tiny 912}} & 0.92$\times$ \\
\midrule
B300 & \raggedright $4{\times}4096$, 16 h, $d{=}128$, causal & \shortstack{0.13\\{\tiny 2048}} & \shortstack{0.17\\{\tiny 1603}} & \shortstack{{\tiny\texttt{no}}\\{\tiny\texttt{causal}}} & \shortstack{0.21\\{\tiny 1277}} & \shortstack{\textbf{0.13}\\{\tiny 2061}} & \shortstack{0.28\\{\tiny 980}} & \shortstack{0.19\\{\tiny 1459}} & \shortstack{0.17\\{\tiny 1626}} & \shortstack{{\tiny\texttt{no}}\\{\tiny\texttt{causal}}} & \shortstack{0.20\\{\tiny 1401}} & \shortstack{0.24\\{\tiny 1129}} & \shortstack{0.21\\{\tiny 1290}} & \shortstack{0.18\\{\tiny 1515}} & \shortstack{0.20\\{\tiny 1388}} & 0.99$\times$ \\
 & \raggedright $4{\times}4096$, 16 h, $d{=}128$, non-causal & \shortstack{0.23\\{\tiny 2420}} & \shortstack{0.31\\{\tiny 1789}} & \shortstack{\textbf{0.20}\\{\tiny 2763}} & \shortstack{0.32\\{\tiny 1706}} & \shortstack{0.21\\{\tiny 2581}} & \shortstack{0.27\\{\tiny 2031}} & \shortstack{0.28\\{\tiny 1976}} & \shortstack{0.23\\{\tiny 2397}} & \shortstack{0.30\\{\tiny 1836}} & \shortstack{0.30\\{\tiny 1849}} & \shortstack{0.33\\{\tiny 1664}} & \shortstack{0.35\\{\tiny 1576}} & \shortstack{0.30\\{\tiny 1846}} & \shortstack{0.30\\{\tiny 1822}} & 0.88$\times$ \\
 & \raggedright $1{\times}32768$, 16 h, $d{=}128$, causal & \shortstack{1.90\\{\tiny 2339}} & \shortstack{2.40\\{\tiny 1832}} & \shortstack{{\tiny\texttt{no}}\\{\tiny\texttt{causal}}} & \shortstack{2.48\\{\tiny 1774}} & \shortstack{\textbf{1.80}\\{\tiny 2444}} & \shortstack{2.52\\{\tiny 1744}} & \shortstack{2.21\\{\tiny 1990}} & \shortstack{1.94\\{\tiny 2265}} & \shortstack{{\tiny\texttt{no}}\\{\tiny\texttt{causal}}} & \shortstack{2.88\\{\tiny 1527}} & \shortstack{3.11\\{\tiny 1415}} & \shortstack{3.02\\{\tiny 1455}} & \shortstack{2.82\\{\tiny 1562}} & \shortstack{2.91\\{\tiny 1510}} & 0.95$\times$ \\
 & \raggedright $1{\times}32768$, 16 h, $d{=}128$, non-causal & \shortstack{3.50\\{\tiny 2515}} & \shortstack{4.66\\{\tiny 1887}} & \shortstack{\textbf{3.01}\\{\tiny 2927}} & \shortstack{4.37\\{\tiny 2011}} & \shortstack{3.06\\{\tiny 2872}} & \shortstack{3.73\\{\tiny 2355}} & \shortstack{3.91\\{\tiny 2248}} & \shortstack{3.11\\{\tiny 2315}} & \shortstack{3.81\\{\tiny 2297}} & \shortstack{4.46\\{\tiny 1999}} & \shortstack{4.97\\{\tiny 1768}} & \shortstack{4.87\\{\tiny 1793}} & \shortstack{4.66\\{\tiny 1911}} & \shortstack{4.50\\{\tiny 1954}} & 0.86$\times$ \\
 & \raggedright $2{\times}8192$, 16 h, $d{=}128$, causal & \shortstack{0.24\\{\tiny 2256}} & \shortstack{0.32\\{\tiny 1712}} & \shortstack{{\tiny\texttt{no}}\\{\tiny\texttt{causal}}} & \shortstack{0.38\\{\tiny 1447}} & \shortstack{\textbf{0.23}\\{\tiny 2431}} & \shortstack{0.46\\{\tiny 1208}} & \shortstack{0.32\\{\tiny 1712}} & \shortstack{0.29\\{\tiny 1914}} & \shortstack{{\tiny\texttt{no}}\\{\tiny\texttt{causal}}} & \shortstack{0.34\\{\tiny 1612}} & \shortstack{0.42\\{\tiny 1301}} & \shortstack{0.38\\{\tiny 1436}} & \shortstack{0.32\\{\tiny 1703}} & \shortstack{0.34\\{\tiny 1594}} & 0.93$\times$ \\
 & \raggedright $2{\times}8192$, 16 h, $d{=}128$, non-causal & \shortstack{0.45\\{\tiny 2426}} & \shortstack{0.60\\{\tiny 1841}} & \shortstack{\textbf{0.38}\\{\tiny 2859}} & \shortstack{0.60\\{\tiny 1825}} & \shortstack{0.40\\{\tiny 2763}} & \shortstack{0.50\\{\tiny 2210}} & \shortstack{0.54\\{\tiny 2052}} & \shortstack{0.41\\{\tiny 2651}} & \shortstack{0.55\\{\tiny 1986}} & \shortstack{0.57\\{\tiny 1917}} & \shortstack{0.64\\{\tiny 1731}} & \shortstack{0.66\\{\tiny 1663}} & \shortstack{0.57\\{\tiny 1921}} & \shortstack{0.58\\{\tiny 1909}} & 0.85$\times$ \\
 & \raggedright $1{\times}16384$, 16 h, $d{=}128$, causal & \shortstack{0.46\\{\tiny 2378}} & \shortstack{0.62\\{\tiny 1777}} & \shortstack{{\tiny\texttt{no}}\\{\tiny\texttt{causal}}} & \shortstack{0.70\\{\tiny 1569}} & \shortstack{\textbf{0.41}\\{\tiny 2661}} & \shortstack{0.81\\{\tiny 1367}} & \shortstack{0.58\\{\tiny 1882}} & \shortstack{0.50\\{\tiny 2208}} & \shortstack{{\tiny\texttt{no}}\\{\tiny\texttt{causal}}} & \shortstack{0.64\\{\tiny 1718}} & \shortstack{0.81\\{\tiny 1358}} & \shortstack{0.73\\{\tiny 1473}} & \shortstack{0.61\\{\tiny 1798}} & \shortstack{0.65\\{\tiny 1695}} & 0.89$\times$ \\
 & \raggedright $1{\times}16384$, 16 h, $d{=}128$, non-causal & \shortstack{0.88\\{\tiny 2494}} & \shortstack{1.17\\{\tiny 1873}} & \shortstack{\textbf{0.76}\\{\tiny 2900}} & \shortstack{1.17\\{\tiny 1873}} & \shortstack{0.80\\{\tiny 2837}} & \shortstack{0.96\\{\tiny 2293}} & \shortstack{1.04\\{\tiny 2142}} & \shortstack{0.84\\{\tiny 2625}} & \shortstack{1.03\\{\tiny 2136}} & \shortstack{1.32\\{\tiny 1722}} & \shortstack{1.46\\{\tiny 1507}} & \shortstack{1.45\\{\tiny 1542}} & \shortstack{1.34\\{\tiny 1643}} & \shortstack{1.36\\{\tiny 1749}} & 0.86$\times$ \\
 & \raggedright $4{\times}4096$, 16/4 h, $d{=}128$, causal & \shortstack{0.13\\{\tiny 2081}} & \shortstack{0.17\\{\tiny 1609}} & \shortstack{{\tiny\texttt{no}}\\{\tiny\texttt{causal}}} & \shortstack{0.21\\{\tiny 1284}} & \shortstack{\textbf{0.13}\\{\tiny 2077}} & \shortstack{0.25\\{\tiny 1082}} & \shortstack{0.18\\{\tiny 1550}} & \shortstack{0.17\\{\tiny 1655}} & \shortstack{{\tiny\texttt{no}}\\{\tiny\texttt{causal}}} & \shortstack{0.18\\{\tiny 1532}} & \shortstack{0.22\\{\tiny 1272}} & \shortstack{0.23\\{\tiny 1341}} & \shortstack{0.17\\{\tiny 1576}} & \shortstack{0.18\\{\tiny 1514}} & 0.99$\times$ \\
 & \raggedright $4{\times}4096$, 16/4 h, $d{=}128$, non-causal & \shortstack{0.23\\{\tiny 2422}} & \shortstack{0.31\\{\tiny 1786}} & \shortstack{{\tiny\texttt{no}}\\{\tiny\texttt{GQA}}} & \shortstack{0.32\\{\tiny 1715}} & \shortstack{\textbf{0.21}\\{\tiny 2626}} & \shortstack{0.27\\{\tiny 2041}} & \shortstack{0.28\\{\tiny 1975}} & \shortstack{0.23\\{\tiny 2411}} & \shortstack{{\tiny\texttt{no}}\\{\tiny\texttt{GQA}}} & \shortstack{0.29\\{\tiny 1878}} & \shortstack{0.33\\{\tiny 1661}} & \shortstack{0.34\\{\tiny 1642}} & \shortstack{0.30\\{\tiny 1859}} & \shortstack{0.30\\{\tiny 1827}} & 0.93$\times$ \\
 & \raggedright $4{\times}4096$, 16 h, $d{=}64$, causal & \shortstack{0.14\\{\tiny 973}} & \shortstack{0.20\\{\tiny 698}} & \shortstack{{\tiny\texttt{no}}\\{\tiny\texttt{causal}}} & \shortstack{0.16\\{\tiny 871}} & \shortstack{\textbf{0.12}\\{\tiny 1134}} & \shortstack{{\tiny\texttt{no}}\\{\tiny\texttt{d=64}}} & \shortstack{0.18\\{\tiny 764}} & \shortstack{0.23\\{\tiny 587}} & \shortstack{{\tiny\texttt{no}}\\{\tiny\texttt{causal}}} & \shortstack{0.14\\{\tiny 968}} & \shortstack{0.23\\{\tiny 588}} & \shortstack{0.25\\{\tiny 553}} & \shortstack{0.14\\{\tiny 985}} & \shortstack{0.15\\{\tiny 888}} & 0.87$\times$ \\
 & \raggedright $4{\times}4096$, 16 h, $d{=}64$, non-causal & \shortstack{0.23\\{\tiny 1211}} & \shortstack{0.28\\{\tiny 967}} & \shortstack{{\tiny\texttt{no}}\\{\tiny\texttt{d=64}}} & \shortstack{0.23\\{\tiny 1169}} & \shortstack{\textbf{0.19}\\{\tiny 1416}} & \shortstack{{\tiny\texttt{no}}\\{\tiny\texttt{d=64}}} & \shortstack{0.23\\{\tiny 1174}} & \shortstack{0.29\\{\tiny 942}} & \shortstack{{\tiny\texttt{no}}\\{\tiny\texttt{d=64}}} & \shortstack{0.23\\{\tiny 1221}} & \shortstack{0.31\\{\tiny 886}} & \shortstack{0.31\\{\tiny 897}} & \shortstack{0.23\\{\tiny 1199}} & \shortstack{0.23\\{\tiny 1178}} & 0.85$\times$ \\
\bottomrule
\end{tabular}
\end{adjustbox}
\end{table}

\begin{table}[!htbp]
\centering
\caption{Backward kernels on batched sequences, kernels alone with operands quantized outside the timed region: time in ms (fastest in bold), TFLOP/s in small type. Speedup: the fastest baseline's time over ours.}\label{tab:batched_bwd}
\scriptsize
\setlength{\tabcolsep}{0.5pt}
\renewcommand{\arraystretch}{1.0}
\begin{adjustbox}{max width=\linewidth,max totalheight=0.87\textheight}
\begin{tabular}{@{}lp{3.6cm}llllllllll@{}}
\toprule
 & & \multicolumn{2}{l}{\textbf{Ours}} & \multicolumn{3}{l}{\textbf{8-bit}} & \multicolumn{4}{l}{\textbf{BF16}} & \\
\cmidrule(lr){3-4}\cmidrule(lr){5-7}\cmidrule(lr){8-11}
\textbf{GPU} & \textbf{Shape} & MXFP8 & NVFP4 & \shortstack{cuDNN\\FP8} & \shortstack{cuDNN\\MX} & LP-FA4 MXFP8 & FA4 & cuDNN & SDPA & LP-FA4 & \textbf{Speedup} \\
\midrule
B200 & \raggedright $4{\times}4096$, 16 h, $d{=}128$, causal & \shortstack{0.53\\{\tiny 1299}} & \shortstack{\textbf{0.51}\\{\tiny 1356}} & \shortstack{0.57\\{\tiny 1213}} & \shortstack{0.64\\{\tiny 1078}} & {\tiny\texttt{no causal}} & \shortstack{0.71\\{\tiny 971}} & \shortstack{0.64\\{\tiny 1076}} & \shortstack{0.75\\{\tiny 922}} & \shortstack{0.70\\{\tiny 985}} & 1.12$\times$ \\
 & \raggedright $4{\times}4096$, 16 h, $d{=}128$, non-causal & \shortstack{0.87\\{\tiny 1575}} & \shortstack{\textbf{0.83}\\{\tiny 1659}} & \shortstack{0.94\\{\tiny 1469}} & \shortstack{1.12\\{\tiny 1224}} & \shortstack{1.14\\{\tiny 1210}} & \shortstack{1.19\\{\tiny 1153}} & \shortstack{1.24\\{\tiny 1111}} & \shortstack{1.29\\{\tiny 1063}} & \shortstack{1.51\\{\tiny 909}} & 1.13$\times$ \\
 & \raggedright $1{\times}32768$, 16 h, $d{=}128$, causal & \shortstack{7.03\\{\tiny 1565}} & \shortstack{\textbf{6.65}\\{\tiny 1654}} & \shortstack{7.95\\{\tiny 1384}} & \shortstack{9.00\\{\tiny 1221}} & {\tiny\texttt{no causal}} & \shortstack{9.28\\{\tiny 1185}} & \shortstack{9.61\\{\tiny 1144}} & \shortstack{9.63\\{\tiny 1141}} & \shortstack{9.15\\{\tiny 1202}} & 1.20$\times$ \\
 & \raggedright $1{\times}32768$, 16 h, $d{=}128$, non-causal & \shortstack{13.6\\{\tiny 1621}} & \shortstack{\textbf{13.0}\\{\tiny 1687}} & \shortstack{14.6\\{\tiny 1510}} & \shortstack{17.6\\{\tiny 1247}} & \shortstack{15.9\\{\tiny 1379}} & \shortstack{17.3\\{\tiny 1270}} & \shortstack{18.6\\{\tiny 1183}} & \shortstack{18.7\\{\tiny 1178}} & \shortstack{22.7\\{\tiny 967}} & 1.12$\times$ \\
 & \raggedright $2{\times}8192$, 16 h, $d{=}128$, causal & \shortstack{0.90\\{\tiny 1523}} & \shortstack{\textbf{0.86}\\{\tiny 1590}} & \shortstack{1.03\\{\tiny 1338}} & \shortstack{1.21\\{\tiny 1136}} & {\tiny\texttt{no causal}} & \shortstack{1.28\\{\tiny 1076}} & \shortstack{1.20\\{\tiny 1150}} & \shortstack{1.31\\{\tiny 1052}} & \shortstack{1.20\\{\tiny 1142}} & 1.19$\times$ \\
 & \raggedright $2{\times}8192$, 16 h, $d{=}128$, non-causal & \shortstack{1.73\\{\tiny 1587}} & \shortstack{\textbf{1.63}\\{\tiny 1682}} & \shortstack{1.86\\{\tiny 1474}} & \shortstack{2.25\\{\tiny 1219}} & \shortstack{2.11\\{\tiny 1305}} & \shortstack{2.25\\{\tiny 1219}} & \shortstack{2.40\\{\tiny 1145}} & \shortstack{2.45\\{\tiny 1124}} & \shortstack{2.91\\{\tiny 944}} & 1.14$\times$ \\
 & \raggedright $1{\times}16384$, 16 h, $d{=}128$, causal & \shortstack{1.79\\{\tiny 1538}} & \shortstack{\textbf{1.67}\\{\tiny 1648}} & \shortstack{2.04\\{\tiny 1350}} & \shortstack{2.32\\{\tiny 1185}} & {\tiny\texttt{no causal}} & \shortstack{2.43\\{\tiny 1131}} & \shortstack{2.43\\{\tiny 1130}} & \shortstack{2.50\\{\tiny 1101}} & \shortstack{2.38\\{\tiny 1153}} & 1.22$\times$ \\
 & \raggedright $1{\times}16384$, 16 h, $d{=}128$, non-causal & \shortstack{3.40\\{\tiny 1617}} & \shortstack{\textbf{3.19}\\{\tiny 1725}} & \shortstack{3.66\\{\tiny 1501}} & \shortstack{4.41\\{\tiny 1246}} & \shortstack{4.09\\{\tiny 1345}} & \shortstack{4.43\\{\tiny 1240}} & \shortstack{4.70\\{\tiny 1171}} & \shortstack{4.82\\{\tiny 1140}} & \shortstack{5.75\\{\tiny 956}} & 1.15$\times$ \\
 & \raggedright $4{\times}4096$, 16/4 h, $d{=}128$, causal & \shortstack{0.50\\{\tiny 1384}} & \shortstack{\textbf{0.47}\\{\tiny 1452}} & \shortstack{0.69\\{\tiny 1000}} & \shortstack{0.75\\{\tiny 918}} & {\tiny\texttt{no causal}} & \shortstack{0.76\\{\tiny 903}} & \shortstack{0.66\\{\tiny 1042}} & \shortstack{0.77\\{\tiny 891}} & \texttt{n.m.} & 1.39$\times$ \\
 & \raggedright $4{\times}4096$, 16/4 h, $d{=}128$, non-causal & \shortstack{0.96\\{\tiny 1426}} & \shortstack{\textbf{0.92}\\{\tiny 1492}} & \shortstack{1.06\\{\tiny 1292}} & \shortstack{1.27\\{\tiny 1085}} & {\tiny\texttt{no GQA}} & \shortstack{1.26\\{\tiny 1093}} & \shortstack{1.26\\{\tiny 1092}} & \shortstack{1.34\\{\tiny 1024}} & \texttt{n.m.} & 1.16$\times$ \\
 & \raggedright $4{\times}4096$, 16 h, $d{=}64$, causal & \shortstack{0.42\\{\tiny 821}} & \shortstack{\textbf{0.41}\\{\tiny 839}} & \shortstack{0.50\\{\tiny 684}} & {\tiny\texttt{no d=64}} & {\tiny\texttt{no causal}} & \shortstack{0.50\\{\tiny 691}} & \shortstack{0.43\\{\tiny 793}} & \shortstack{0.49\\{\tiny 697}} & \shortstack{0.61\\{\tiny 562}} & 1.06$\times$ \\
 & \raggedright $4{\times}4096$, 16 h, $d{=}64$, non-causal & \shortstack{0.73\\{\tiny 943}} & \shortstack{\textbf{0.71}\\{\tiny 970}} & \shortstack{0.86\\{\tiny 801}} & {\tiny\texttt{no d=64}} & {\tiny\texttt{no d=64}} & \shortstack{0.82\\{\tiny 840}} & \shortstack{0.80\\{\tiny 858}} & \shortstack{0.89\\{\tiny 775}} & \shortstack{1.03\\{\tiny 667}} & 1.13$\times$ \\
\midrule
B300 & \raggedright $4{\times}4096$, 16 h, $d{=}128$, causal & \shortstack{0.52\\{\tiny 1313}} & \shortstack{0.50\\{\tiny 1386}} & \shortstack{\textbf{0.46}\\{\tiny 1496}} & \shortstack{0.57\\{\tiny 1199}} & {\tiny\texttt{no causal}} & \shortstack{0.64\\{\tiny 1080}} & \shortstack{0.59\\{\tiny 1169}} & \shortstack{0.59\\{\tiny 1167}} & \shortstack{0.77\\{\tiny 891}} & 0.93$\times$ \\
 & \raggedright $4{\times}4096$, 16 h, $d{=}128$, non-causal & \shortstack{0.84\\{\tiny 1626}} & \shortstack{0.80\\{\tiny 1716}} & \shortstack{\textbf{0.73}\\{\tiny 1869}} & \shortstack{0.94\\{\tiny 1467}} & \shortstack{1.05\\{\tiny 1309}} & \shortstack{1.02\\{\tiny 1347}} & \shortstack{1.03\\{\tiny 1326}} & \shortstack{1.04\\{\tiny 1324}} & \shortstack{1.35\\{\tiny 1024}} & 0.92$\times$ \\
 & \raggedright $1{\times}32768$, 16 h, $d{=}128$, causal & \shortstack{6.43\\{\tiny 1709}} & \shortstack{\textbf{6.20}\\{\tiny 1773}} & \shortstack{6.48\\{\tiny 1743}} & \shortstack{7.97\\{\tiny 1380}} & {\tiny\texttt{no causal}} & \shortstack{8.69\\{\tiny 1266}} & \shortstack{9.21\\{\tiny 1193}} & \shortstack{9.27\\{\tiny 1185}} & \shortstack{10.2\\{\tiny 1083}} & 1.05$\times$ \\
 & \raggedright $1{\times}32768$, 16 h, $d{=}128$, non-causal & \shortstack{11.8\\{\tiny 1863}} & \shortstack{11.4\\{\tiny 1936}} & \shortstack{\textbf{10.7}\\{\tiny 2067}} & \shortstack{13.5\\{\tiny 1646}} & \shortstack{12.9\\{\tiny 1707}} & \shortstack{13.9\\{\tiny 1586}} & \shortstack{14.9\\{\tiny 1471}} & \shortstack{15.0\\{\tiny 1467}} & \shortstack{19.6\\{\tiny 1068}} & 0.94$\times$ \\
 & \raggedright $2{\times}8192$, 16 h, $d{=}128$, causal & \shortstack{0.90\\{\tiny 1524}} & \shortstack{0.85\\{\tiny 1614}} & \shortstack{\textbf{0.79}\\{\tiny 1734}} & \shortstack{1.00\\{\tiny 1380}} & {\tiny\texttt{no causal}} & \shortstack{1.05\\{\tiny 1305}} & \shortstack{1.03\\{\tiny 1332}} & \shortstack{1.03\\{\tiny 1328}} & \shortstack{1.31\\{\tiny 1053}} & 0.92$\times$ \\
 & \raggedright $2{\times}8192$, 16 h, $d{=}128$, non-causal & \shortstack{1.59\\{\tiny 1733}} & \shortstack{1.50\\{\tiny 1837}} & \shortstack{\textbf{1.37}\\{\tiny 2001}} & \shortstack{1.76\\{\tiny 1563}} & \shortstack{1.85\\{\tiny 1488}} & \shortstack{1.86\\{\tiny 1483}} & \shortstack{1.96\\{\tiny 1409}} & \shortstack{1.96\\{\tiny 1405}} & \shortstack{2.52\\{\tiny 1093}} & 0.92$\times$ \\
 & \raggedright $1{\times}16384$, 16 h, $d{=}128$, causal & \shortstack{1.70\\{\tiny 1616}} & \shortstack{\textbf{1.59}\\{\tiny 1737}} & \shortstack{1.67\\{\tiny 1644}} & \shortstack{2.02\\{\tiny 1361}} & {\tiny\texttt{no causal}} & \shortstack{2.21\\{\tiny 1248}} & \shortstack{2.32\\{\tiny 1187}} & \shortstack{2.32\\{\tiny 1191}} & \shortstack{2.58\\{\tiny 1069}} & 1.05$\times$ \\
 & \raggedright $1{\times}16384$, 16 h, $d{=}128$, non-causal & \shortstack{3.16\\{\tiny 1740}} & \shortstack{\textbf{2.96}\\{\tiny 1855}} & \shortstack{3.06\\{\tiny 1795}} & \shortstack{3.76\\{\tiny 1461}} & \shortstack{3.67\\{\tiny 1499}} & \shortstack{4.18\\{\tiny 1314}} & \shortstack{4.43\\{\tiny 1239}} & \shortstack{4.44\\{\tiny 1239}} & \shortstack{5.30\\{\tiny 1042}} & 1.03$\times$ \\
 & \raggedright $4{\times}4096$, 16/4 h, $d{=}128$, causal & \shortstack{0.48\\{\tiny 1428}} & \shortstack{\textbf{0.47}\\{\tiny 1467}} & \shortstack{0.57\\{\tiny 1204}} & \shortstack{0.69\\{\tiny 1004}} & {\tiny\texttt{no causal}} & \shortstack{0.68\\{\tiny 1006}} & \shortstack{0.62\\{\tiny 1123}} & \shortstack{0.61\\{\tiny 1118}} & \shortstack{0.82\\{\tiny 838}} & 1.22$\times$ \\
 & \raggedright $4{\times}4096$, 16/4 h, $d{=}128$, non-causal & \shortstack{0.91\\{\tiny 1508}} & \shortstack{0.89\\{\tiny 1541}} & \shortstack{\textbf{0.87}\\{\tiny 1582}} & \shortstack{1.06\\{\tiny 1273}} & {\tiny\texttt{no GQA}} & \shortstack{1.17\\{\tiny 1266}} & \shortstack{1.19\\{\tiny 1178}} & \shortstack{1.16\\{\tiny 1187}} & \shortstack{1.41\\{\tiny 977}} & 0.97$\times$ \\
 & \raggedright $4{\times}4096$, 16 h, $d{=}64$, causal & \shortstack{0.40\\{\tiny 865}} & \shortstack{\textbf{0.38}\\{\tiny 902}} & \shortstack{0.41\\{\tiny 846}} & {\tiny\texttt{no d=64}} & {\tiny\texttt{no causal}} & \shortstack{0.47\\{\tiny 728}} & \shortstack{0.42\\{\tiny 809}} & \shortstack{0.47\\{\tiny 734}} & \shortstack{0.55\\{\tiny 628}} & 1.07$\times$ \\
 & \raggedright $4{\times}4096$, 16 h, $d{=}64$, non-causal & \shortstack{0.67\\{\tiny 1025}} & \shortstack{\textbf{0.65}\\{\tiny 1065}} & \shortstack{0.67\\{\tiny 1027}} & {\tiny\texttt{no d=64}} & {\tiny\texttt{no d=64}} & \shortstack{0.78\\{\tiny 884}} & \shortstack{0.79\\{\tiny 871}} & \shortstack{0.83\\{\tiny 832}} & \shortstack{0.91\\{\tiny 752}} & 1.04$\times$ \\
\bottomrule
\end{tabular}
\end{adjustbox}
\end{table}

\end{document}